%% file: main.tex
\documentclass[10pt,twocolumn,letterpaper]{article}

\usepackage[pagenumbers]{cvpr} 

\input{preamble}

\definecolor{cvprblue}{rgb}{0.21,0.49,0.74}
\usepackage[pagebackref,breaklinks,colorlinks,allcolors=cvprblue]{hyperref}

\def\confName{CVPR}
\def\confYear{2027}

\title{PRISM: Profiling and Representation of Individual Stereotypical Motion}

\author{
Austin A. MBaye\\
Northeastern University\\
Boston, MA, USA\\
{\tt\small mbaye.au@northeastern.edu}
\and
Savik Kinger\\
Yale University\\
New Haven, CT, USA\\
{\tt\small savik.kinger@yale.edu}
\and
Matthew S. Goodwin\\
Northeastern University\\
Boston, MA, USA\\
{\tt\small m.goodwin@northeastern.edu}
}

\begin{document}
\maketitle
\input{sec/0_abstract}    
\input{sec/1_intro}

\input{sec/3_method}
\input{sec/3b_properties}

\input{sec/4_experiments}
\input{sec/5_results}
\input{sec/6_discussion}
{
    \small
    \bibliographystyle{ieeenat_fullname}
    \bibliography{main}
}

\input{sec/X_suppl}

\end{document}

%% file: preamble.tex
\usepackage{microtype}
\usepackage{amsmath,amssymb,amsfonts}
\usepackage{booktabs}
\usepackage{multirow}
\usepackage{graphicx}
\usepackage{xspace}
\usepackage{microtype}
\usepackage{enumitem}
\usepackage{mathtools}
\usepackage{amsthm}

\usepackage{tabularx}
\usepackage{array}

\renewcommand{\paragraph}[1]{\vspace{.5em}\noindent\textbf{#1.}}

\newtheorem{proposition}{Proposition}
\newtheorem{example}{Example}

\newcommand{\SW}{S\mathbb{W}}
\newcommand{\R}{\mathbb{R}}

%% file: sec/0_abstract.tex
\begin{abstract}
An anatomical pose provides interpretable joint structure, but no fixed set of physical landmarks can fully specify a person's movement. We introduce PRISM, a complementary representation that characterizes repetitive movement using tracked masks and dense optical flow. Torso-relative flow measures intensity, frame-wise mask variation measures spatial extent, signed-flow recurrence estimates frequency, and persistent sliding-window geometry describes the rhythmicity of the movement. Across 1,890 synthetic conditions, the measurements recover controlled speed, amplitude, and frequency while exposing observation limits. An idealized hand-motion experiment demonstrates variation omitted by identical exact COCO-17 trajectories; controlled corruption experiments separately examine measurement sensitivity. Audits of two pose methods on the same data cohort and 36,930 external skeleton clips document incomplete observation and dependence on processing choices. In 1,404 annotated complex stereotypical motor movement bouts from six autistic participants, total flow converges with pose motion, while matched-coordinate examples and repeated-session profiles show differences in intensity, extent, frequency, and rhythmicity within behavior labels. A separate VLM comparison evaluates quantitative frequency estimation, and a profile-conditioned decoder links the measurements to illustrative skeletal movements. PRISM connects numerical profiles to visible examples, providing an interpretable way to compare how the same behavior is expressed beyond its label or event count.
\end{abstract}

%% file: sec/1_intro.tex
\section{Introduction}
\label{sec:intro}

Developing approaches to understand human behavior has long been a goal of computer vision. The challenges faced in accomplishing this arise, in part, from the inherent variability of behaviors, as documented in clinical contexts \cite{Melo2020Prevalence, Huang_2023_CVPR, Amraee_2026_CVPR, Popoola2012, Robertson2005}. For example, a behavior categorized as "body rocking'' may look completely different for two different people, or the same person on two different days. Abstract and subjective notes make it difficult for clinicians to interpret and reach consensus on the significance of varying movements. This challenge is intensified by a shortage of comprehensive, annotated datasets capturing real-world behavioral diversity, particularly among clinical groups \cite{Rajagopalan2013SSBD, Huang_2023_CVPR}. In particular, there is a lack of standardized metrics for quantifying inherent variability of movements in the same behavior "category.''

In autism, for example, detecting high-impact behaviors—including self-injury (biting, head banging), aggression (hitting, pushing), and repetitive motor movements (body rocking, hand flapping) is essential for diagnosis, functional analysis of behavior, assessing maturational development, and evaluating response to intervention. Designing robust behavioral detection systems has therefore been a focus of much recent work, particularly because newer deep-learning-based vision models are strong at handling diverse clinical presentations and behavioral complexity \cite{Amraee_2026_CVPR, li2026deterministicevidencelayervisionlanguage}. While these approaches have shown promise in classifying behavior, reliably detecting and quantifying the underlying intensity and variability of these behaviors remains an open challenge.

In this paper, we focus on a commonly studied class of complex stereotypical motor movements (SMMs), specifically body rocking and hand or arm flapping in autistic individuals \cite{DSM5,LaGrow1984}. Automated approaches identify these behaviors using wearable sensors, pose, and video models \cite{Albinali2009,Goodwin2011Automated,wei2023vision,negin2021vision,AQSM2026}. However, two correctly labeled bouts can differ substantially on movement qualities such as intensity, repetition rate, and temporal regularity. Describing these differences requires measurements of how movement is expressed, with explicit scales and sensitivity to recording conditions.

An important choice is the observation space in which these differences are measured. Video analyses commonly summarize movement using anatomical landmarks \cite{AQSM2026, Barami2024}, such as hand and shoulder locations, or learned features. H-MoRe \cite{huang2025hmorelearninghumancentricmotion} highlights that skeletal pose omits body-shape information important for tasks such as gait recognition, motivating a human-centric representation that preserves both motion and shape. For SMM characterization, this surfaces a related question: which differences in visible movement does the chosen representation retain? A fixed landmark configuration can leave hand shape and motion between landmarks unspecified, allowing visibly different movements to share the same observed landmark trajectories. Pose estimation introduces additional uncertainty through localization errors, missing observations, and inconsistent tracking across frames. Moreover, model-specific parameters to tune, such as visibility thresholds and tracking confidences, greatly influence outcomes. We therefore distinguish information omitted by the landmark configuration from errors in estimating its coordinates. 

A dense motion field preserves spatial detail, but comparing fields across behavioral bouts is difficult. A useful movement description should answer four complementary questions: how fast does the person move relative to a central-body reference, how much does their silhouette vary in space, how quickly does the movement repeat, and how regular are the repetitive movements? These dimensions describe aspects of a behavior that a categorical label or event count leaves unspecified. Keeping them allows similar movement speeds to accompany different spatial extents, repetition rates, or regularity. A regional, observation-based representation lets us examine these differences without requiring consistent localization of individual joints. 

\begin{figure*}[htb]
    \centering
    \includegraphics[width=.65\textwidth]{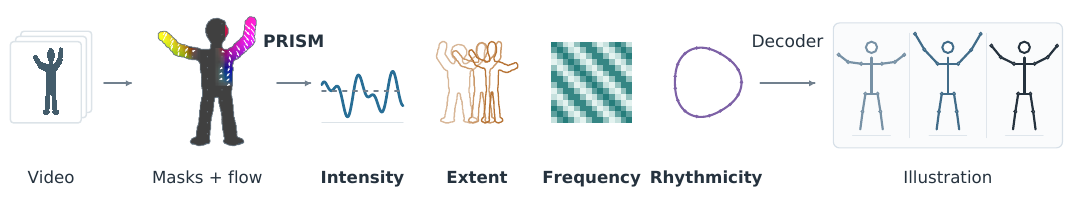}
    \caption{\textbf{PRISM overview.} The tracked mask defines target-person support. Torso-relative flow yields intensity; camera-corrected total signed flow yields recurrence frequency and sliding-window periodicity. Common-grid masks independently yield frame-wise RMS extent. Event boundaries yield duration and between-bout timing.}
    \label{fig:overview}
\end{figure*}

We introduce PRISM, a tracked-region framework for characterizing relative movement intensity and variability within identified SMM bouts. PRISM is designed for settings with only monocular video, without additional sensors. Starting from continuously tracked person masks \cite{Videnovic2025DAM4SAM}, we construct camera-corrected dense-flow states and measure frame-wise silhouette variation. These observations define a compact profile of \emph{intensity}, \emph{extent}, \emph{frequency}, and \emph{rhythmicity} \ref{fig:overview}. We additionally show the sufficiency of these metrics for the task of measuring variability, and provide a system by which clinicians can interpret the dynamics of these metrics while preserving individual privacy.

Our contributions include: (i) an interpretable framework for characterizing SMM from monocular video, (ii) a sparse-observation formulation that identifies movement omitted by a specified skeleton landmark configuration, (iii) a link between numerical profiles and visible movement for interpretability, (iv) empirical evaluation through synthetic experiments and characterization of 1,404 SMM bouts, along with a quantitative VLM comparison. Together, these contributions establish a regional measurement framework that complements anatomical pose representations and enables finer-grained analyses of how an identified behavior is expressed over time and space.

%% file: sec/3_method.tex
\section{Method}
\label{sec:method}
Given a tracked mask $M_t\in\{0,1\}^{H\times W}$ and an identified bout $b$, PRISM returns:
\begin{equation}
    \Phi(b)=(I_b,E_b,\widehat f_b,PS_b).
\end{equation}
The coordinates measure typical torso-relative speed, silhouette variation, recurrence rate, and rhythmicity\footnote{Bout boundaries are provided; behavior detection is outside the scope of this study.}. 

\subsection{Tracked-Region Motion}
The person is tracked through the complete recording before masks are restricted to bout intervals \cite{Ravi2024SAM2,Videnovic2025DAM4SAM}. In real video, background correspondences outside a dilated mask support a RANSAC similarity transform allowing translation, in-plane rotation, and isotropic scaling. Synthetic camera motion is removed using the known simulated transform. DIS computes camera-corrected displacement $v_t(p)$ between adjacent frames \cite{Kroeger2016DIS}. The retained support $\mathcal A_t$ requires mask-consistent origins, displaced endpoints, and forward--backward consistency.

Using smoothed mask height $\widetilde h_t$ and frame interval $\Delta t_t$, normalized signed flow is:
\begin{equation}
    F_t(p)=\frac{v_t(p)}{\widetilde h_t\Delta t_t},\qquad p\in\mathcal A_t.
    \label{eq:normalized-flow}
\end{equation}
Horizontal and vertical components retain movement direction; replacing them with magnitude alone would identify opposing movement phases.

\subsection{Torso-Relative Intensity}
Bulk body translation can dominate visible motion. We estimate a reference velocity as the component-wise median within a compact elliptical neighborhood $C_t\subseteq\mathcal A_t$ of the mask bounding-box midpoint:
\begin{equation}
    u_t=\operatorname{median}_{p\in C_t}F_t(p), \qquad \widetilde F_t(p)=F_t(p)-u_t.
\end{equation}
The core is a torso proxy rather than an anatomical estimate. For the represented upright and seated postures, reference subtraction suppresses central-body translation while preserving motion relative to that core. Bout intensity is:
\begin{equation}
    I_b=\operatorname{median}_{t\in b}\operatorname Q_{.75}\{\|\widetilde F_t(p)\|_2:p\in\mathcal A_t\}.
    \label{eq:intensity}
\end{equation}
The spatial upper quartile reduces dependence on isolated extreme vectors, and the temporal median describes typical bout motion. Units are approximately tracked body heights/second. Total-flow intensity and torso-translation speed are retained separately.

\subsection{Frame-wise Spatial Extent}
Let $N_b$ be the number of retained binary masks on the common fixed crop and $A_b=\operatorname{median}_t|M_t|$ their median area. With $x_t=\operatorname{vec}(M_t)/\sqrt{A_b}$ and $\bar x_b=N_b^{-1}\sum_t x_t$, extent is:
\begin{equation}
    E_b=\left[\frac1{N_b}\sum_{t\in b}\|x_t-\bar x_b\|_2^2\right]^{1/2}.
    \label{eq:extent}
\end{equation}
This RMS orbit radius measures area-normalized silhouette variation. 

\subsection{Recurrence Frequency and Cyclic Organization}
\label{sec:frequency}
Unlike intensity, recurrence uses camera-corrected \emph{total} signed flow. Each valid-person bounding crop is resampled to a $48\times32$ grid $\Omega_G$. After capping extreme magnitudes using a bout-specific median-absolute-deviation threshold, the state is:
\begin{equation}
    z_t=\operatorname{vec}\left(\sqrt{q_t(p)}F_t(p)\right)_{p\in\Omega_G},
    \label{eq:flow-state}
\end{equation}
where $q_t(p)\in[0,1]$ weights valid flow support. We retain the longest consecutive valid flow run. This trajectory summarizes directional image motion across the normalized person region.

\paragraph{Frequency}
For $n$ retained states, define:
\begin{equation}
    D^F_{ij}=\frac{\|z_i-z_j\|_2}{\sqrt{\max(\operatorname{median}_t\|z_t\|_2^2,\epsilon)}}, \qquad \epsilon>0.
    \label{eq:flow-distance}
\end{equation}
The mean distance at lag $\ell$ is:
\begin{equation}
    r_F(\ell)=\frac1{n-\ell}\sum_{i=1}^{n-\ell}D^F_{i,i+\ell}.
    \label{eq:flow-lag-profile}
\end{equation}
Recurrent directional motion produces a repeating lag profile. Its dominant spectral peak gives $\widehat f_b$, with quadratic interpolation for off-grid precision.

\paragraph{Rhythmicity}
Frequency describes recurrence rate, but does not specify how rhythmic it is. We construct a sliding-window embedding of the signed-flow trajectory. Let $z_t$ denote the linearly interpolated flow-state trajectory. Given embedding dimension $d$ and delay $\tau$, define:
\begin{equation}
    p_t = \SW_{d,\tau}z(t) = \left(z(t), z(t+\tau_b), \ldots, z(t+d\tau_b) \right).
    \label{eq:flow-sw}
\end{equation}
Following the periodic sliding-window construction \cite{PereaHarer2015}, we choose the window length using the estimated period $\hat L = 1/\widehat f_b$ so that $d\tau=(d/(d+1))\hat L.$ We fix $d=23$ across bouts, consistent with the oscillatory complexity observed in the cohort and a prior analysis \cite{AQSM2026}. 

The resulting collection of windows $X_b = \left\{ \SW_{d,\tau}z(t)\right\}_t$ forms a point cloud in the space of short directional motion sequences. Each window is centered by $\bar z_t=(d+1)^{-1}\sum_{k=0}^{d}z(t+k\tau_b)$, giving $C(p_t)=(z(t)-\bar z_t,\ldots,z(t+d\tau_b)-\bar z_t)$. Nonzero windows are sphere-normalized:
\begin{equation}
    \widetilde X_b=\{C(p)/\|C(p)\|_2:p\in X_b,\ \|C(p)\|_2>0\}.
    \label{eq:flow-normalization}
\end{equation}
Centering removes mean directional motion and normalization removes uniform window-magnitude scaling. A periodic (or rhythmic) movement generates an approximately closed loop in this space because windows separated by one cycle contain similar sequences of motion fields.

We then compute the first Vietoris--Rips persistence diagram using Ripser.py \cite{Tralie2018Ripser,Bauer2021Ripser}. For finite classes ordered by decreasing persistence $m_i=d_i-b_i$,
\begin{equation}
    PS_b=e^{-b_1/2}\frac{m_1}{\sqrt3}-e^{-b_2/2}\frac{m_2}{\sqrt3}.
    \label{eq:ps}
\end{equation}
An absent class contributes zero. The score rewards one dominant loop and penalizes a competing loop. It describes cyclic organization under this representation and embedding, rather than timing consistency alone. Low scores can accompany inconsistent cycle geometry, competing rhythms, or degraded observations; unavailable measurements remain distinct from numerical scores.

%% file: sec/3b_properties.tex
\section{Relation to Pose and Measurement Rationale}
\label{sec:rationale}
PRISM makes two choices: it observes movement over a tracked person region using dense optical flow, and describes that movement with several complementary measurements. The following results explain the rationale for each choice. First, we establish when a specified pose-landmark observation retains enough information to determine a movement property. Second, we show why intensity, extent, and frequency resolve different movement factors even in a simple oscillator. Proofs are provided in the supplement.

\subsection{Sparse Landmark Observation}

H-MoRe motivates human-centric dense motion representations that retain body shape and fine motion information beyond skeletal landmarks \cite{huang2025hmorelearninghumancentricmotion}. We formalize a restricted question: when does a fixed landmark-velocity observation preserve a particular movement measurement? We first separate information omitted by a specified landmark configuration from errors in estimating it. At one frame, let $F\in\R^{2P}$ be a dense field and $S:\R^{2P}\to\R^{2J}$ a fixed linear landmark-velocity observation, such as point sampling or local averaging, with $J<P$.

\begin{proposition}[Recoverability from sparse observations]
    \label{prop:sparse-main}
    The map satisfies $\dim\ker S\geq2(P-J)>0$. A functional $Q:\mathcal A\to\R$ on an admissible class $\mathcal A$ can be written as $Q=q\circ S$ precisely when:
    \begin{equation}
        SF=SG\ \Longrightarrow\ Q(F)=Q(G),\qquad F,G\in\mathcal A.
        \label{eq:fibers}
    \end{equation}
\end{proposition}

For example, a configuration without finger landmarks can omit fine finger movement even when the observed body landmarks are exact. A property that changes with this omitted motion cannot be determined from those landmarks alone. Improving landmark estimation does not resolve that representational ambiguity. Thus, sparse observations recover a particular movement property only when that property is shared by all admissible fields with the same observations. A nontrivial ambient kernel alone does not establish ambiguity within every physical movement class. This model concerns sampled velocities, not arbitrary RGB pose estimation or recovery of anatomical identities from flow. In real video, we separately examine pose availability, processing sensitivity, and convergence with regional measurements.

\subsection{Complementarity in a Restricted Oscillator Model}
Regional observations still require choosing measurements. The next result gives a concrete reason to retain intensity, extent, and frequency separately. Even with exact observations, any two measurements may conceal differences in the generating movement.

Consider a stationary region of area $B>0$ and a disjoint translating rectangle of width $w$ and height $\ell$, with $X(t)=A\sin(2\pi ft+\phi)$. Assume exact instantaneous masks and velocities, a stationary intensity core, known reference height $h$, uniform full-cycle phase sampling, $0<2A<w$, and moving-area fraction $w\ell/(B+w\ell)>1/4$.

\begin{proposition}[Conditional three-factor identification]
    \label{prop:identification-main}
    With exact frequency $F=f$, $c=\sqrt2\pi/h$, and $k=8/\pi^2$, the population measurements satisfy
    \begin{equation}
        I=cAf,\qquad E^2=\frac{kA\ell}{B+w\ell},\qquad F=f.
        \label{eq:identification}
    \end{equation}
    For known $B,w,h$, the triple uniquely identifies $(A,f,\ell)$ on this domain; each coordinate pair leaves a nonconstant local family of generating parameters unresolved.
\end{proposition}

In this model, intensity and frequency determine the oscillation amplitude but leave the moving region's height undetermined. Adding extent resolves this difference. Each of the other measurement pairs also leaves some movement differences unresolved, giving a concrete reason to retain all three under these assumptions. The sinusoidal model does not vary the regularity of repetition, so we examine what $PS$ adds through separate controls that disrupt timing while keeping intensity, extent, and frequency matched.

%% file: sec/4_experiments.tex
\section{Experimental Design}
\label{sec:experiments}

\subsection{Synthetic construct recovery.} We evaluate PRISM of 1,890 synthetic video test cases generated with ''Skeleton Lab,'' which renders body movements with known generating parameters. Arm flapping, body rocking, and hand flapping correspond to the movements studied in our cohort \cite{Albinali2009,Goodwin2014,AQSM2026}. The head banging and head shaking presets provide additional tests of localized head motion, broadening the geometric validation beyond the cohort's behavior categories. For each movement, we combine five frequencies, five durations, three amplitudes, and three random seeds, giving 1,125 unperturbed clips. These use a fixed camera and simulator-generated masks, with no added image or signal degradations. We compare intensity with projected speed, extent with primary angular excursion, and frequency with the generating rate. For intensity and extent, the references use different units, so we assess whether the measurements follow changes in the intended properties. For frequency, we report error distributions and the fraction of conditions yielding an estimate. Another 765 cases test sensitivity to changes in video quality, masks, and camera conditions. We compare each perturbed case with its matched baseline and use known camera transforms to separate measurement performance from stabilization error.

\subsection{Matched and rendered organization controls.} We first ask whether rhythmicity can change while intensity, extent, and estimated frequency remain fixed. Thirty-two exact Eulerian conditions combine eight phase seeds with four modulation strengths. We adjust amplitude, moving-region height, and frequency to match the three baseline measurements before computing persistence. Twenty-four compensation controls retain these adjustments but remove modulation. We then test the image-based pipeline using rendered body rocking with stochastic timing disruption or a competing rhythm, evaluating $PS$ at both estimated and reference periods. These rendered controls allow the other measurements to vary. Shared and numerically repeated conditions leave 13 distinct bundles across 24 design records.

\subsection{Real-video profiling.} We analyze 1,404 expert-annotated bouts (616 flap, 600 rock, 188 flap--rock) from six autistic participants aged 12--20 years, repeatedly recorded in laboratory and classroom settings at approximately 4--6~fps \cite{Albinali2009,Goodwin2014,AQSM2026}. The bout tables span 26 source files. We extract bouts from person tracks using the fixed expert boundaries. Duration and preceding inter-bout interval (IBI) are calculated within each recording session. IBI runs from the previous annotated bout's offset to the current bout's onset, regardless of behavior; first bouts and negative intervals are treated as unavailable. We compare total pose motion estimated within padded SAM-mask crops with total flow on exactly shared valid frame pairs. We report pooled and behavior-specific associations, together with correlations of ranks within participant--behavior groups. Confidence intervals use 2,000 participant-clustered bootstrap replicates. Matched bouts and repeated-session profiles describe variation within the recordings; participant, acquisition conditions, and setting are not experimentally separated.

\subsection{Pose audits.} We examine AQSM MediaPipe pose \cite{AQSM2026} and SAM-crop YOLOv11 pose on the same 1,404 bouts, alongside 36,930 external ASDMotion clips \cite{Barami2024}. At a confidence threshold of $.5$, we measure landmark coverage and flag gaps and temporal or skeletal discontinuities in unsmoothed trajectories. We also vary confidence thresholds, short-gap interpolation, and smoothing to assess their effect on motion summaries. Confidence scores and coordinate systems differ across sources, so we interpret thresholds within each system. The flags identify sequences for review and are not verified pose errors. ASDMotion review is restricted to skeletons. Full diagnostic definitions and sensitivity analyses are in the supplement.

\subsection{Comparison with vision--language models.} We compare direct VLM estimates with PRISM on 40 synthetic clips combining five behaviors, generating frequencies of 1 and 2 Hz, amplitude multipliers of $.6$ and $1.4$, and durations of 3 and 5 s. VLMs receive timestamped RGB frames at 10 fps without demonstrations. PRISM uses the same sampling rate and additionally receives tracked masks. We report frequency mean absolute error over answered clips, the number of valid answers, and the fraction of all 40 clips estimated within $20\%$ of the generating frequency. Missing or invalid responses count as failures in this last measure. Extended movement-ordering comparisons and a separate 30-fps PRISM condition are in the supplement.

\subsection{Profile-conditioned movement.} We train the same 451-parameter network on nested libraries of 144, 576, and 1,440 clips to predict simulator settings from the four measurements and behavior indicators. We evaluate each library on the same 54 held-out targets across body rocking, arm flapping, and hand flapping. Training excludes the exact test parameter values and generating seeds. We render predicted settings with new seeds and remeasure the resulting movements. We compare the neural decoder with four-coordinate retrieval, retrieval without $PS$, and random same-behavior examples, using IQR scales fixed from the original 144-library profiles. The main illustration uses the 1,440-clip model; the full sample-size experiment is in the supplement.

\subsection{Human subjects.} The original recordings were collected under Groden Center and University of Rhode Island review-board approval with parental informed consent. This study analyzes de-identified recordings. Full protocols, multi-view analyses, extended VLM comparisons, and observation qualifications are provided in the supplement.

%% file: sec/5_results.tex
\section{Results}
\label{sec:results}

\subsection{Construct Recovery and Observation Limits}

On the clean benchmark, intensity and extent generally followed changes in their corresponding generating properties (Table~\ref{tab:construct}). Across geometries with defined correlations, the median Spearman correlation was $.953$ for intensity and $.958$ for extent. Both associations were weaker for head shaking, showing that recovery depends on movement geometry. The hand-flapping speed reference was constant across the design, leaving its intensity correlation undefined.

\begin{table}[htb]
    \small
    \centering
    \caption{\textbf{Clean construct recovery.} Spearman correlations across 225 conditions per geometry. Intensity is compared with projected speed, extent with angular excursion, and frequency with generating rate.}
    \label{tab:construct}
    \begin{tabular}{lrrr}
        \toprule
        Geometry & $I\;\rho$ & $E\;\rho$ & $\widehat f\;\rho$ \\
        \midrule
        Arm flapping & 0.973 & 0.969 & 0.962 \\
        Body rocking & 0.933 & 0.974 & 0.970 \\
        Hand flapping & --- & 0.950 & 0.960 \\
        Head banging & 0.984 & 0.958 & 0.966 \\
        Head shaking & 0.699 & 0.858 & 0.980 \\
        \bottomrule
    \end{tabular}
\end{table}

Signed-flow frequency had a pooled median relative error of $.230\%$ and a mean error of $4.38\%$, with $91.73\%$ of estimates within $20\%$ of the generating rate. Errors were concentrated in short clips. Success within $20\%$ rose from $72.0\%$ at 2 s to $86.7\%$ at 3 s and $100\%$ at each of 5, 7.5, and 10 s, respectively. Grouping by observed cycles showed the same pattern: success was $31.11\%$ among the 135 conditions containing fewer than 2.5 true cycles and $100\%$ among the remaining 990. The tested grid jumps from 2.25 to 3 cycles, so it does not locate a sharp minimum-cycle threshold.

We computed intensity, extent, and flow frequency for all 1,890 conditions, and $PS$ for 1,860, including 1,095 clean conditions. The perturbation tests identify where these measurements became unreliable. At RGB-noise standard deviation 5, the median relative frequency change from the matched clean result was $85.9\%$. Mask errors strongly affected extent: median relative changes reached $279.7\%$ under .5-pixel boundary jitter and $535.4\%$ under $5\%$ partial mask occlusion. Small clean extent values amplify these ratios; absolute changes and the full perturbation results are in the supplement. Measurements were unchanged under correction with the known synthetic camera transforms. This control isolates downstream measurement performance from errors in estimating camera motion.

\subsection{Rhythmicity at Matched Intensity, Extent, and Frequency}
\label{sec:matched-results}

The exact-field controls show that rhythmicity can change while intensity, extent, and estimated frequency remain matched. Across all 32 Eulerian conditions, intensity and extent matched the baseline within each seed to relative error below $3\times10^{-16}$, and estimated frequency to below $2\times10^{-10}$. Mean $PS$ nevertheless decreased from $.949$ to $.788$, $.672$, and $.514$, respectively, as phase modulation increased (Table~\ref{tab:matched}). All 24 irregular conditions scored below their matched baselines, with scores decreasing at every modulation level in all eight seeds. An independent timing measure showed the same trend: cycle-duration CV increased from zero to approximately $.034$, $.068$, and $.135$, respectively.

\begin{table}[htb]
    \small
    \centering
    \caption{\textbf{Matched Eulerian controls.} Eight seeds per modulation level. Intensity, extent, and estimated frequency are matched before persistence is computed. Measurements use exact fields, isolating the descriptors from optical-flow estimation error.}
    \label{tab:matched}
    \begin{tabular}{rcc}\toprule
        Phase modulation $\varepsilon$ & Mean $PS$ & Mean cycle CV\\\midrule
        0 & .949 & .000\\
        .15 & .788 & .034\\
        .30 & .672 & .068\\
        .60 & .514 & .135\\\bottomrule
    \end{tabular}
\end{table}

Removing modulation while retaining the adjusted geometry and frequency restored high scores. The 24 compensation controls had $PS$ values between $.931$ and $.949$, each exceeding its irregular counterpart. The score reduction therefore follows the timing modulation rather than the compensating parameter changes alone. Together, these controls show that $PS$ distinguishes temporal differences that the other three summaries leave unresolved in this design.

The same trend survived rendering and DIS estimation. With increasing stochastic timing disruption, median $PS$ decreased from $.884$ to $.702$, $.520$, and $.383$, respectively, using estimated periods; the corresponding reference-period values were $.883$, $.688$, $.532$, and $.347$, respectively. Adding a competing rhythm also reduced $PS$, but the strongest mixture shifted the selected frequency toward the added $\sqrt2$-Hz component. At that level, the estimated-period and reference-period scores differed substantially ($.393$ versus $.156$). These rendered examples demonstrate both sensitivity to timing changes and dependence on the rhythm used to set the sliding window. Full rendered-control results are in the supplement.

\subsection{Real-Video Movement Profiles}
\label{sec:real-results}

Among the 1,404 real SMM bouts, intensity was available for 1,399, extent for 1,403, and $PS$ for 762. Of 780 positive frequency estimates, 715 passed the primary eligibility criteria. Requiring at least 2.5 estimated cycles retained 604 bouts and shifted the flapping median from $.641$ to $.789$ Hz, illustrating how cycle-count filtering changes both the sample and its frequency distribution. Median $PS$ was $.032$ for flap, $.131$ for flap--rock, and $.352$ for rock. Nonnegative IBIs were available for 1,378 bouts; duration, IBI, and repeated-session distributions are in the supplement. These distributions provide descriptive summaries of the recorded bouts.

Pose and flow measurements of total motion were strongly associated on shared valid frame pairs. Spearman correlation was $.817$ across all bouts and $.764$ after ranking within participant--behavior groups ($n=1,379$). For articulated pose motion and torso-corrected flow, the corresponding correlations were $.695$ and $.657$ ($n=1,376$). These associations support convergence between the two motion summaries.

\begin{figure}[htb]
    \centering
    \includegraphics[width=\columnwidth]{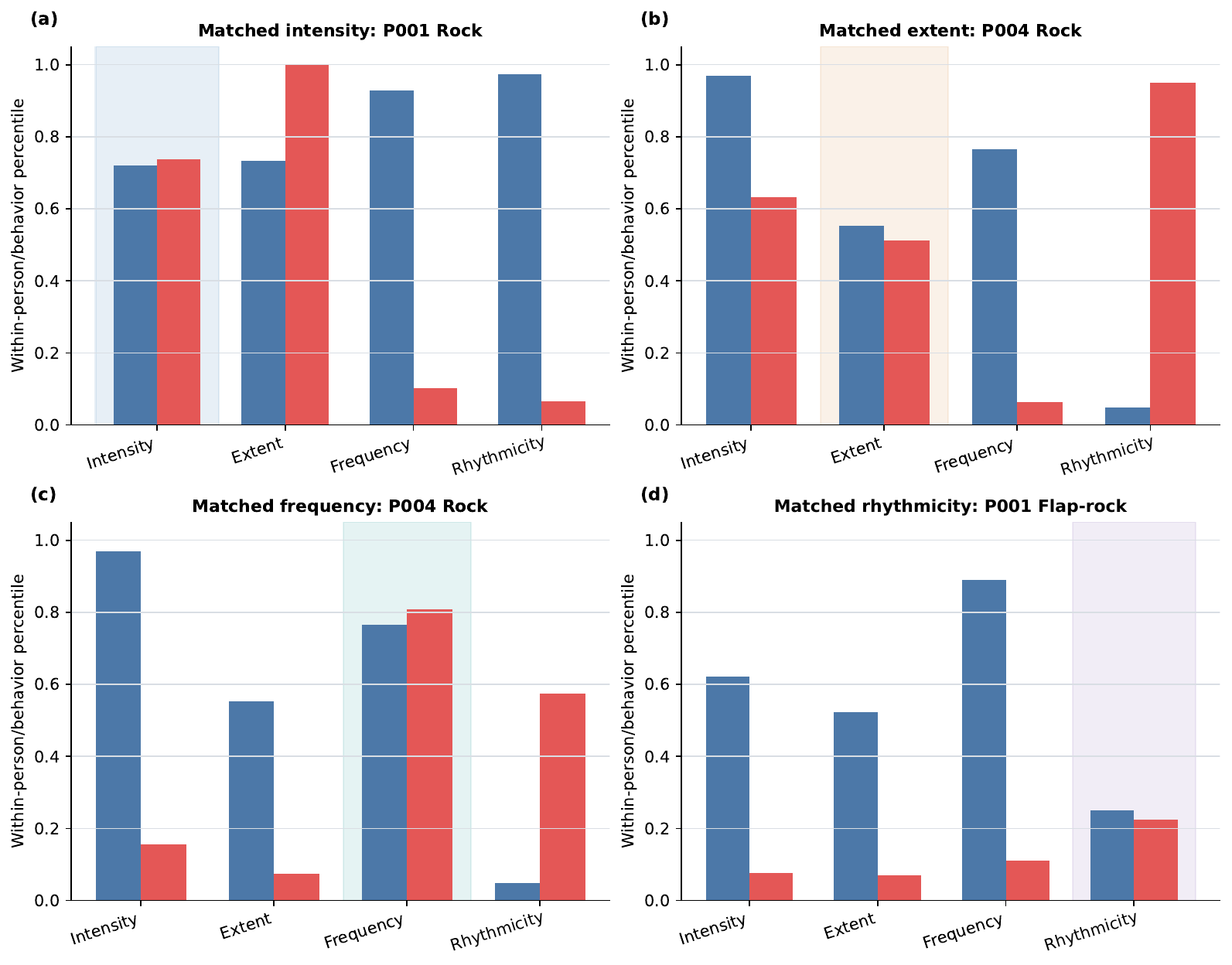}
    \caption{\textbf{Complementary real-video measurements.}
    Selected bouts share participant, behavior, and source recording. The shaded coordinate is matched within six percentile points, while other coordinate ranks can differ. Matching one measurement does not establish equivalent underlying movement or exclude observation error.}
    \label{fig:matched}
\end{figure}

Examples from a single recording illustrate what the four-coordinate profile adds to a behavior label. Two P001 rocking bouts had similar intensity (.211 and .213 body heights/second), but differed in estimated frequency (1.091 and .635~Hz) and cyclic organization ($PS=.592$ and $.020$). Conversely, two P004 bouts had similar estimated frequencies (.848 and .874~Hz), but differed in intensity (.360 and .100 body heights/second; Fig.~\ref{fig:matched}). These examples illustrate complementary measured properties.

\begin{figure*}[htb]
    \centering
    \includegraphics[width=.8\textwidth]{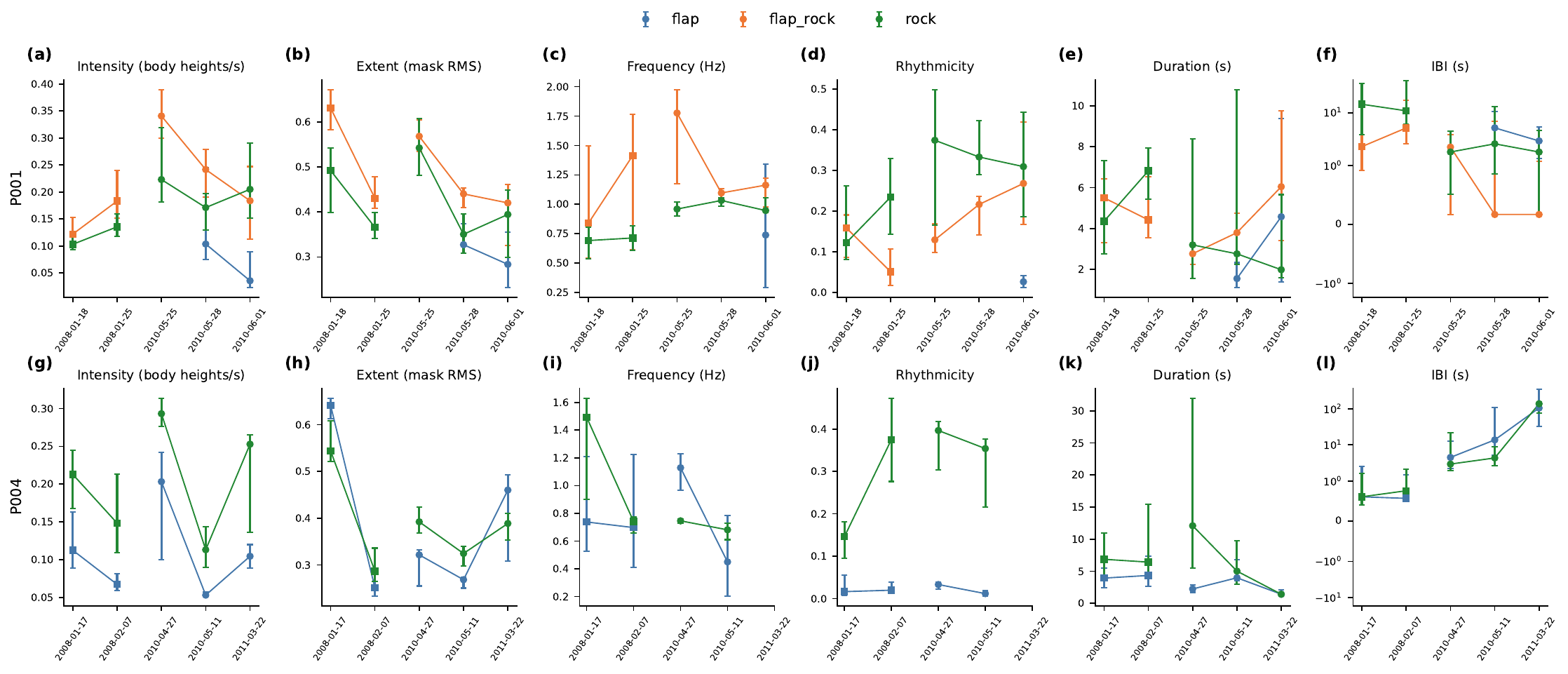}
    \caption{\textbf{Repeated-session movement profiles.}
    Points show behavior-specific medians and bars show bout-level
    interquartile ranges for Participants 1 and 4. Squares and circles
    distinguish studies; lines connect recordings within a study.
    Each summary requires at least three available bouts for that
    coordinate. Same-day recordings have separate run labels.
    The profiles are descriptive and do not establish a
    population-level longitudinal effect.}
    \label{fig:real-profiles}
\end{figure*}
Repeated-session profiles illustrate how the same measurements summarize recordings across time (Fig.~\ref{fig:real-profiles}). For Participants 1 and 4, we report behavior-specific medians and bout-level interquartile ranges separately for each recording. These profiles describe recordings collected under differing acquisition conditions.

\subsection{Pose Availability and Processing Sensitivity}
\label{sec:pose-audit-results}

Pose coverage was incomplete in all three sources. At confidence threshold $.5$, median landmark coverage was $73.8\%$ for AQSM MediaPipe, $64.2\%$ for SAM-crop pose, and $51.4\%$ for the external ASDMotion clips. Measuring motion requires consecutive observations; median valid frame-pair coverage was $73.3\%$, $60.8\%$, and $49.9\%$, respectively. These values describe each source at its own confidence threshold, with different confidence scales and recording conditions. We report joint-level diagnostics and the effects of confidence filtering, interpolation, and smoothing in the supplement. Discontinuity flags remain candidates for review because unusual changes can arise from true motion, camera effects, or estimation errors.

\subsection{Comparison with Direct VLM Measurement}
\label{sec:vlm-results}

At 10-fps sampling, PRISM returned frequency estimates for all 40 clips, with mean absolute error $.178$ Hz and $85\%$ of estimates within $20\%$ of the generating rate (Table~\ref{tab:vlm-frequency}). The tested VLM configurations achieved $7.5$--$47.5\%$ success over all clips. Answer coverage mattered: the configuration with $.393$-Hz MAE returned valid answers for only 14 of 40 clips. Reporting both error and all-clip success keeps those missing answers visible. The comparison uses matched frame sampling, with tracked masks supplied additionally to PRISM. Extended ordering results and the separate 30-fps PRISM condition are in the supplement.

\begin{table}[htb]
    \small
    \centering
    \setlength{\tabcolsep}{3pt}
    \caption{\textbf{Frequency measurement at 10 fps.} MAE is computed over valid answers; success is the percentage of all 40 clips within $20\%$ of the generating frequency. PRISM additionally receives tracked masks.}
    \label{tab:vlm-frequency}
    \begin{tabular}{lrrr}
        \toprule
        Method & MAE (Hz) & Answered & Success \\
        \midrule
        PRISM & $\mathbf{0.178}$ & 40/40 & $\mathbf{85.0\%}$ \\
        Grok 4.20 & 2.225 & 40/40 & $7.5\%$ \\
        Grok 4.3 (low) & 0.393 & 14/40 & $17.5\%$ \\
        Grok 4.7 (low) & 0.419 & 37/40 & $47.5\%$ \\
        Qwen3-VL 2B (4-bit) & 0.617 & 40/40 & $40.0\%$ \\
        \bottomrule
    \end{tabular}
\end{table}

\subsection{From Profiles to Illustrative Movement}
\label{sec:exemplar-results}

With 1,440 library clips, the neural decoder produced outputs that could be rendered and remeasured for all 54 held-out targets. Median scaled profile error was $.087$ for the decoder and $.068$ for four-coordinate retrieval. Neural frequency MAE was $.114$ Hz and $PS$ MAE was $.070$, with no predictions requiring clipping to the library parameter ranges. Training reached the preset limit of 3,000 iterations. Increasing the library from 144 to 1,440 clips reduced median neural error by $55.7\%$ relative to the freshly trained 144-sample baseline, although frequency error was lowest at the intermediate library size. Retrieval gave closer overall profile matches at the largest size, using the same distance for selection and evaluation. Full comparisons and error distributions are in the supplement.

Supplementary Video 3 shows two held-out rocking targets used throughout the sample-size experiment. The larger model more closely reproduced their frequency difference and the direction of their $PS$ difference; labels show the measurements achieved by the rendered outputs. Rendering makes the profile concrete by supplying a body, viewpoint, and motion family. Those choices come from the simulator, and many movements can share the same profile.

%% file: sec/6_discussion.tex
\section{Discussion}
\label{sec:discussion}

PRISM provides a measurement framework for describing how an identified repetitive motor behavior is expressed in monocular video. Intensity, silhouette extent, frequency, and repetition regularity capture complementary aspects of movement. A label such as rocking can cover movements that differ substantially in these properties, as earlier kinematic studies of stereotypic body rocking have demonstrated \cite{Newell01061999}. Measuring tracked regions provides a compact description of movement without requiring consistent localization of individual joints. Duration and inter-bout interval add context about when the behavior occurs and how long it lasts.

The controlled experiments demonstrate why keeping the measurements separate is useful. In the matched controls, timing changes altered $PS$ while intensity, extent, and estimated frequency remained matched. The real-video examples illustrate the same motivation: bouts with comparable intensity can differ in frequency or regularity. These pairs were selected to illustrate contrasts; establishing how common those differences are and their clinical relevance is beyond the scope of the current investigation and planned future work. Agreement between pose and flow also suggests that they capture related movement variation. However, both are estimates from the same video, and an independent reference would be needed to assess their accuracy.

The framework also provides a basis for examining how the choice of motion representation affects the measurements. Our implementation uses DIS optical flow \cite{Kroeger2016DIS}. Human-centric methods such as H-MoRe, which incorporate body shape and describe absolute and relative movement \cite{huang2025hmorelearninghumancentricmotion}, may be better suited to capturing subtle movement differences\footnote{We could not access the repository linked in the H-MoRe paper (\url{https://github.com/haku-huang/h-more}; last checked 13 November 2026), so H-MoRe was not evaluated in this study.}. Future work could test whether these representations improve robustness and sensitivity to variability, while checking that the measurements retain their intended meaning and remain comparable across recordings.

Stereotypy can vary with environmental conditions, and similar movement forms can have different behavioral functions \cite{Hall01112003,Kennedy2000Functions}. Paired with records of context and consequences, PRISM could add detail to functional analyses by describing how movement expression changes across experimental conditions. The controlled tests also show how image quality, segmentation, viewpoint, and bout duration affect the measurements. Comparisons across sessions will require comparable acquisition conditions and repeatability testing to distinguish movement changes from recording effects. Larger studies can build on the six-participant evaluation to examine generalization across people and settings.

The neural decoder gives a numerical profile a visible reference. Conditioned on the measurements and behavior label, it produces simulator settings for a canonical skeletal movement. Re-measurement checks how closely the generated example reproduces the requested profile. The simulator supplies the body and motion family, so each example illustrates one compatible movement rather than reconstructing the participant's anatomy. Profiles and generated examples could make selected movement properties available for collaborative or crowd-sourced analysis without distributing participant footage. 

Repeated recordings under comparable conditions could help document maturational change and response to intervention, addressing the need for consistent quantitative measures identified in earlier developmental research \cite{Symons2005Development,Gilchrist2018Automated}. Differences in repetitive behavior across genetic syndromes \cite{moss2009prevalence} also motivate testing a common measurement framework across diagnostic groups. Pairing movement profiles with physiological or neural recordings could help connect movement expression with underlying neurobiology or pathophysiology \cite{RAPP2005548}. Future studies should test whether these profiles improve sensitivity to associations beyond behavior labels, event counts, and existing clinical scales. These applications will require repeatable measurements and evidence that measured changes correspond to differences that matter to participants and clinicians.

\section*{LLM Usage Statement}
OpenAI Codex was used to assist with code implementation and debugging, literature review, and figure generation. Its use was limited to supporting these tasks. All mathematical analyses, experimental procedures, numerical results, and scientific conclusions were independently reviewed and validated by the authors, who retain full responsibility for the scientific content of this work.

%% file: sec/X_suppl.tex
\clearpage
\setcounter{page}{1}
\maketitlesupplementary

\section{Implementation and Synthetic Benchmark}
\label{app:implementation}
Tracking is performed over complete recordings before restricting masks to annotated bouts. The real-video workflow fits a RANSAC similarity transform using background correspondences outside a dilated target mask and computes DIS flow. Synthetic camera transforms are known and corrected using the simulator transform. Retained vectors must originate inside the current mask, have displaced endpoints inside the next mask, and pass forward--backward consistency. Missing observations are not zero movement.

Intensity subtracts a component-wise flow median in the central elliptical reference core. The temporal state instead uses camera-corrected total signed flow on a body-centered $48\times32$ grid, square-root validity weights, a six-MAD magnitude winsorization threshold, and the longest consecutive valid run. 

\paragraph{Synthetic conditions and deviations.}
The final clean grid crosses five geometries, frequencies $.75,1,2,3,4$ Hz, durations $2,3,5,7.5,10$ s, amplitude scales
$.6,1,1.4$, and three seeds, giving 1,125 conditions. The robustness set contains 765 conditions. Perturbation families include changes in
sampling, blur, RGB noise, JPEG compression, boundary jitter, missing masks, partial occlusion, morphology, and camera observation. For an available measurement $Y$, matched-clean relative deviation is:
\begin{equation}
 \Delta_Y(q)=\frac{|Y_q-Y_{\mathrm{clean}}|}
                  {\max(|Y_{\mathrm{clean}}|,10^{-12})},
\end{equation}
where the clean comparator shares generating geometry, frequency, amplitude, duration, and seed. Missing measurements are reported as
unavailable, not assigned zero deviation. Large relative deviations can occur near a clean value of zero.

\begin{figure*}[htb]\centering
    \includegraphics[width=.9\textwidth]{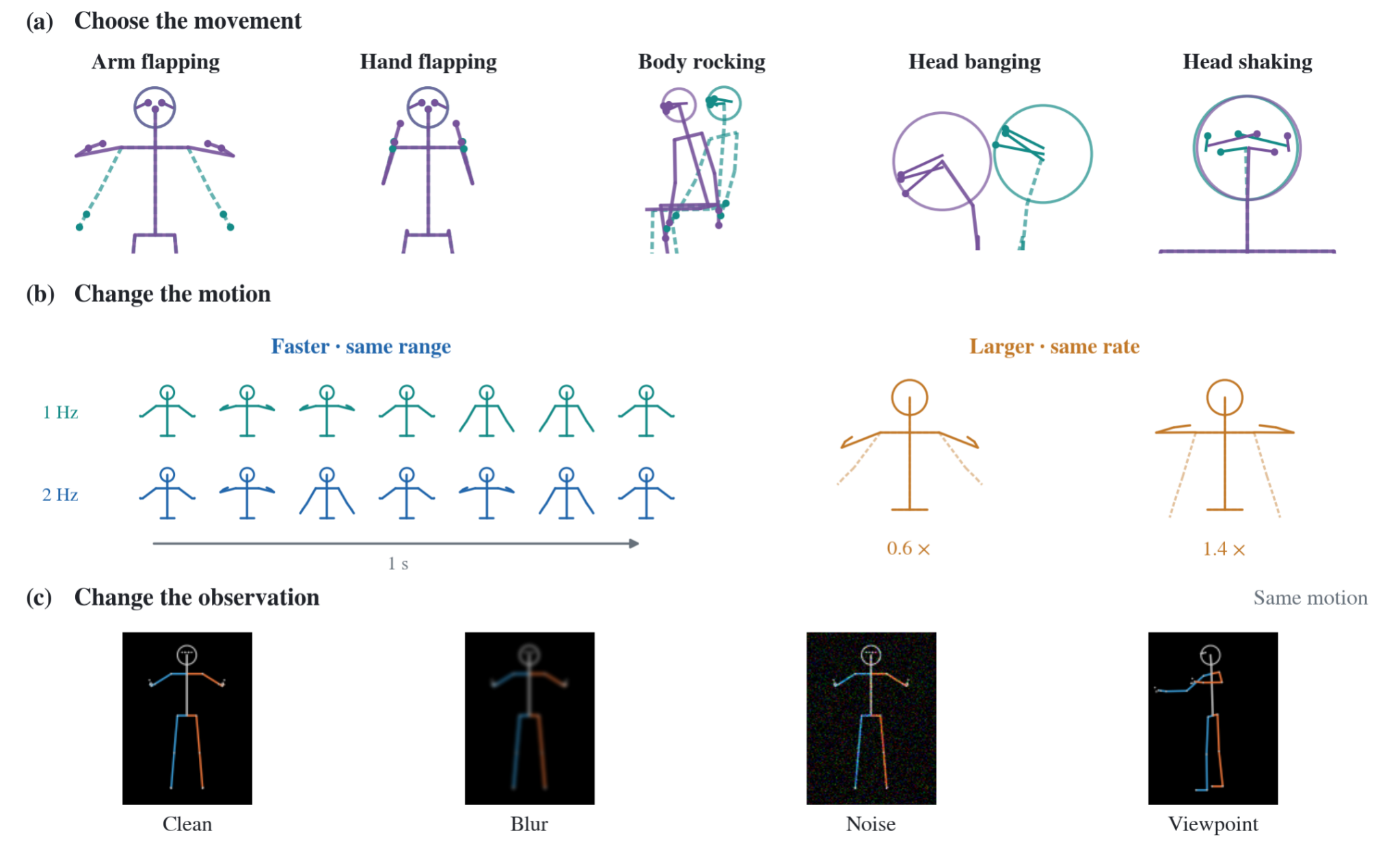}
    \caption{\textbf{Skeleton Lab controls.} Five repetitive geometries vary generating movement properties. Observation perturbations test measurement sensitivity under the benchmark conditions.}
    \label{fig:skeleton-lab-overview}
\end{figure*}

\begin{figure*}[htb]\centering
    \includegraphics[width=.9\textwidth]{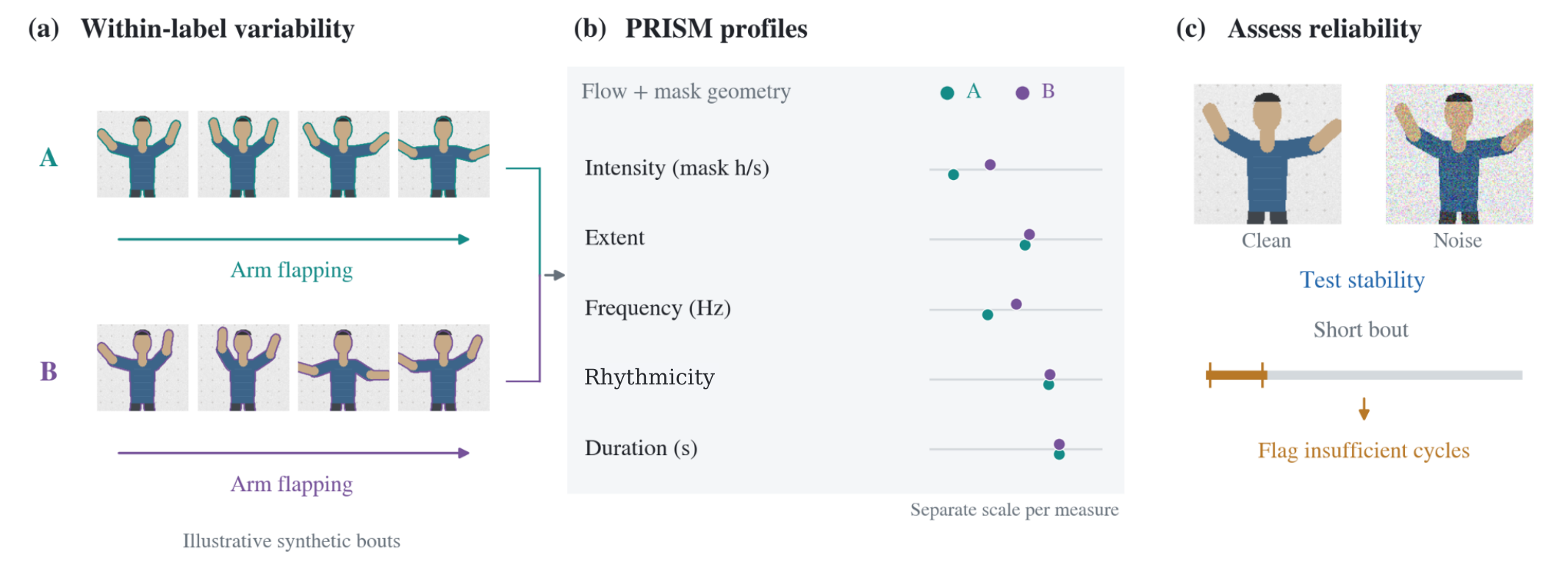}
    \caption{\textbf{Construct and availability logic.} Movements sharing a behavior label can differ in their expression. The framework reports complementary measurements together with availability; unavailable estimates are not interpreted as zero movement or zero organization.}
    \label{fig:variability-overview}
\end{figure*}

\begin{figure*}[p]
    \centering
    \captionsetup[subfigure]{font=small,skip=2pt}
    \begin{subfigure}{\linewidth}
        \centering
        \includegraphics[width=\linewidth,height=.22\textheight,keepaspectratio]{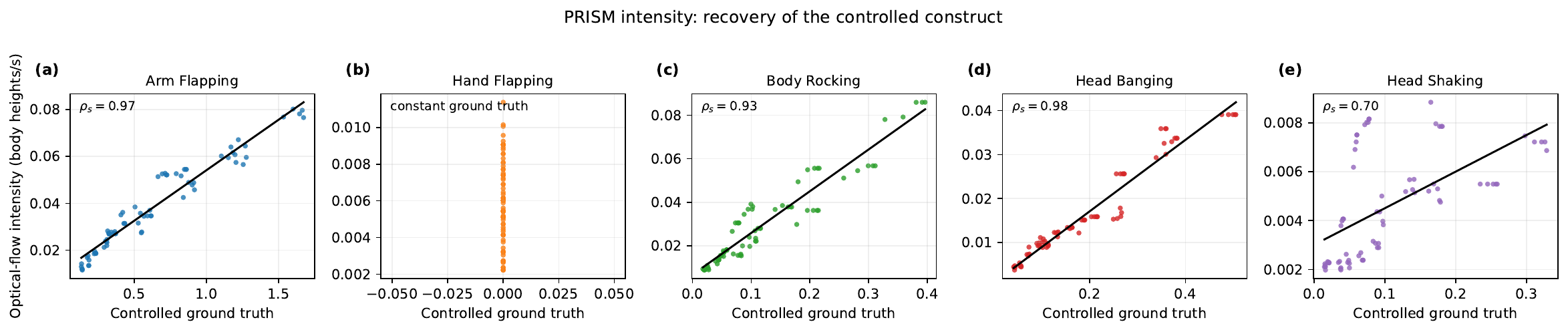}
        \caption{Intensity compared with projected movement speed.}
        \label{fig:intensity_recovery}
    \end{subfigure}
    \par\medskip
    \begin{subfigure}{\linewidth}
        \centering
        \includegraphics[width=\linewidth,height=.22\textheight,keepaspectratio]{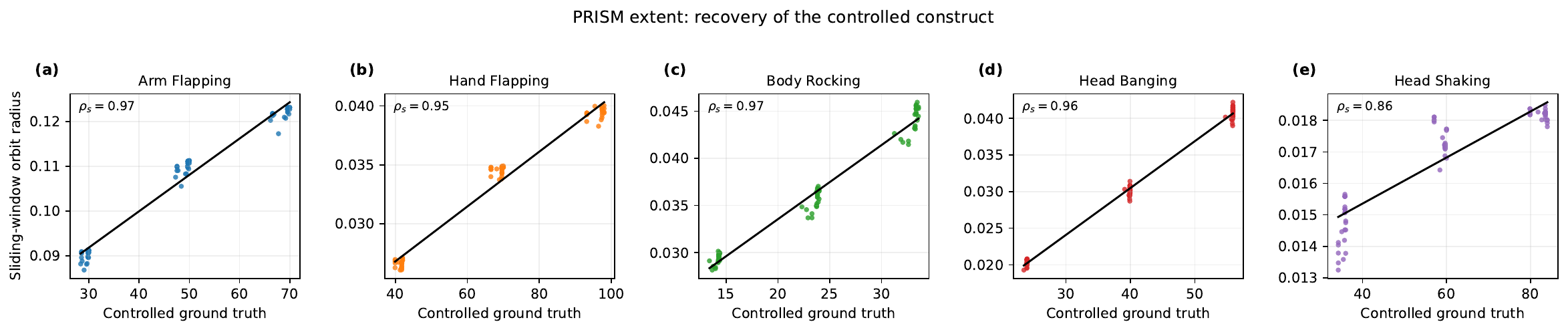}
        \caption{Silhouette extent compared with primary angular excursion.}
        \label{fig:extent_recovery}
    \end{subfigure}
    \par\medskip
    \begin{subfigure}{\linewidth}
        \centering
        \includegraphics[width=\linewidth,height=.22\textheight,keepaspectratio]{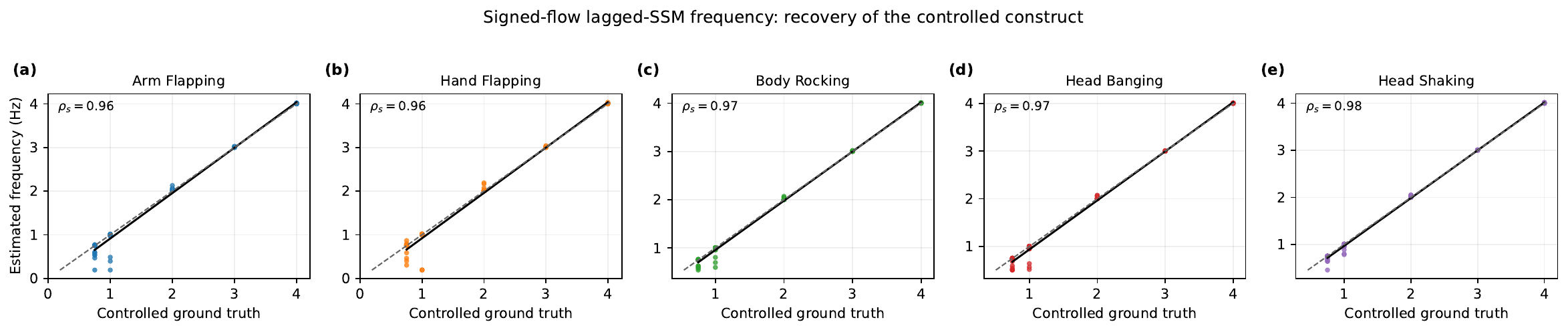}
        \caption{Estimated frequency compared with the generating frequency.}
        \label{fig:freq_recovery}
    \end{subfigure}
    \caption{\textbf{Construct recovery across movement geometries.} The unperturbed benchmark contains 1,125 clips spanning five movement geometries, five frequencies, five durations, three amplitudes, and three seeds. Intensity and extent are compared with references in different units, so their agreement concerns whether they follow changes in speed and excursion rather than numerical equality. Estimated and generating frequencies are both measured in Hz. These comparisons assess recovery within the tested geometries and recording conditions.}
    \label{fig:construct-recovery}
\end{figure*}

\clearpage

\begin{figure*}[p]
    \centering
    \captionsetup[subfigure]{font=small,skip=5pt}
    \begin{subfigure}{0.7\linewidth}
        \centering
        \includegraphics[width=\linewidth]{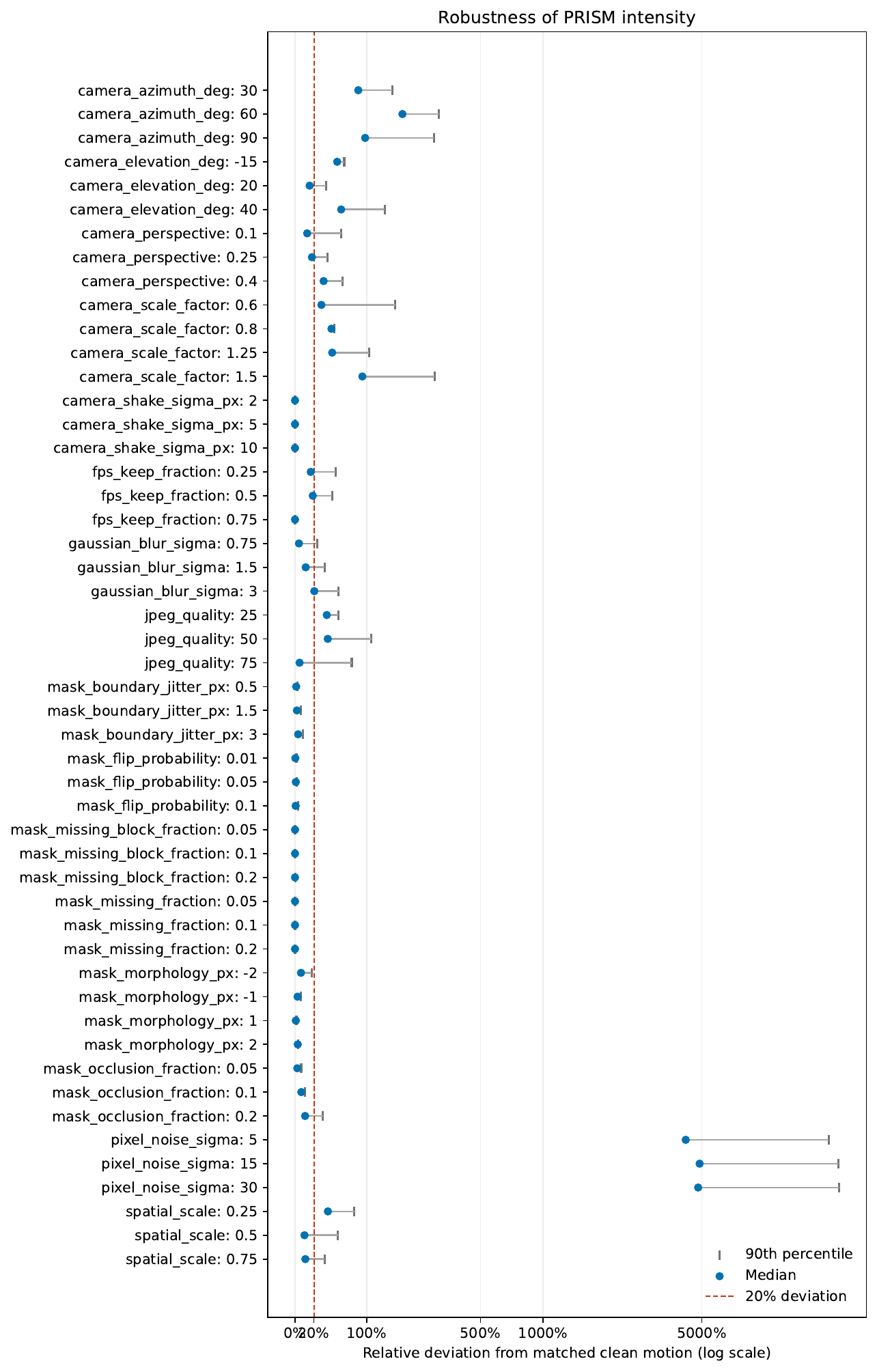}
        \caption{Intensity.}
        \label{fig:intensity_robust}
    \end{subfigure}
    \caption{\textbf{Sensitivity to observation perturbations.} Perturbed clips are compared with matched unperturbed counterparts to examine how observation changes affect each PRISM measurement. Camera-motion conditions use the known synthetic transforms for correction; they therefore test downstream measurements under exact correction rather than the accuracy of estimated camera stabilization.}
    \label{fig:observation-robustness}
\end{figure*}

\clearpage

\begin{figure*}[p]
    \ContinuedFloat
    \centering
    \captionsetup[subfigure]{font=small,skip=5pt}
    \begin{subfigure}{0.7\linewidth}
        \centering
        \includegraphics[width=\linewidth]{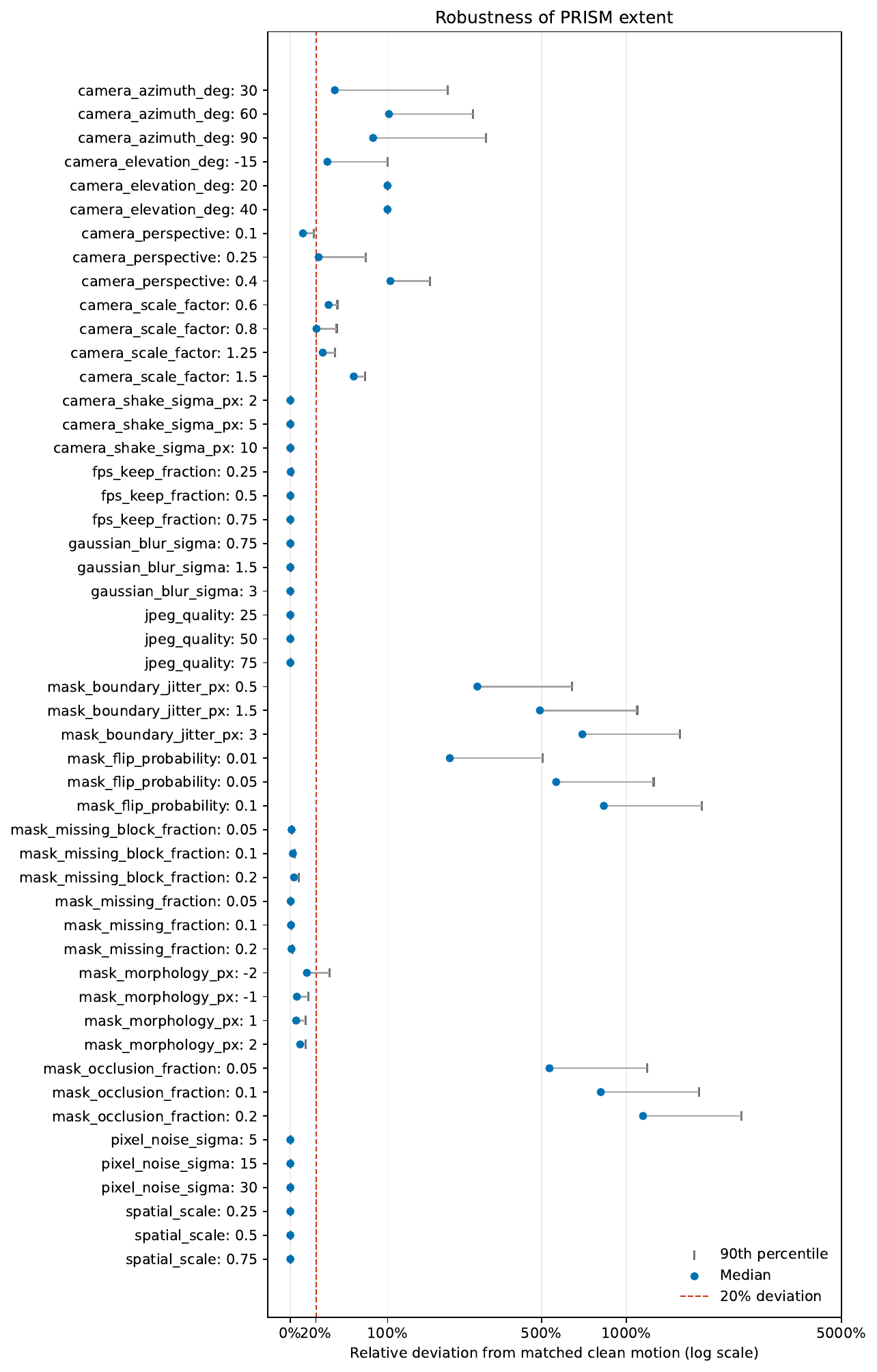}
        \caption{Silhouette extent.}
        \label{fig:extent_robust}
    \end{subfigure}
    \caption[]{\textbf{Sensitivity to observation perturbations, continued.} Silhouette extent under the tested observation perturbations, compared with matched unperturbed counterparts.}
\end{figure*}

\clearpage

\begin{figure*}[p]
    \ContinuedFloat
    \centering
    \captionsetup[subfigure]{font=small,skip=5pt}
    \begin{subfigure}{0.7\linewidth}
        \centering
        \includegraphics[width=\linewidth]{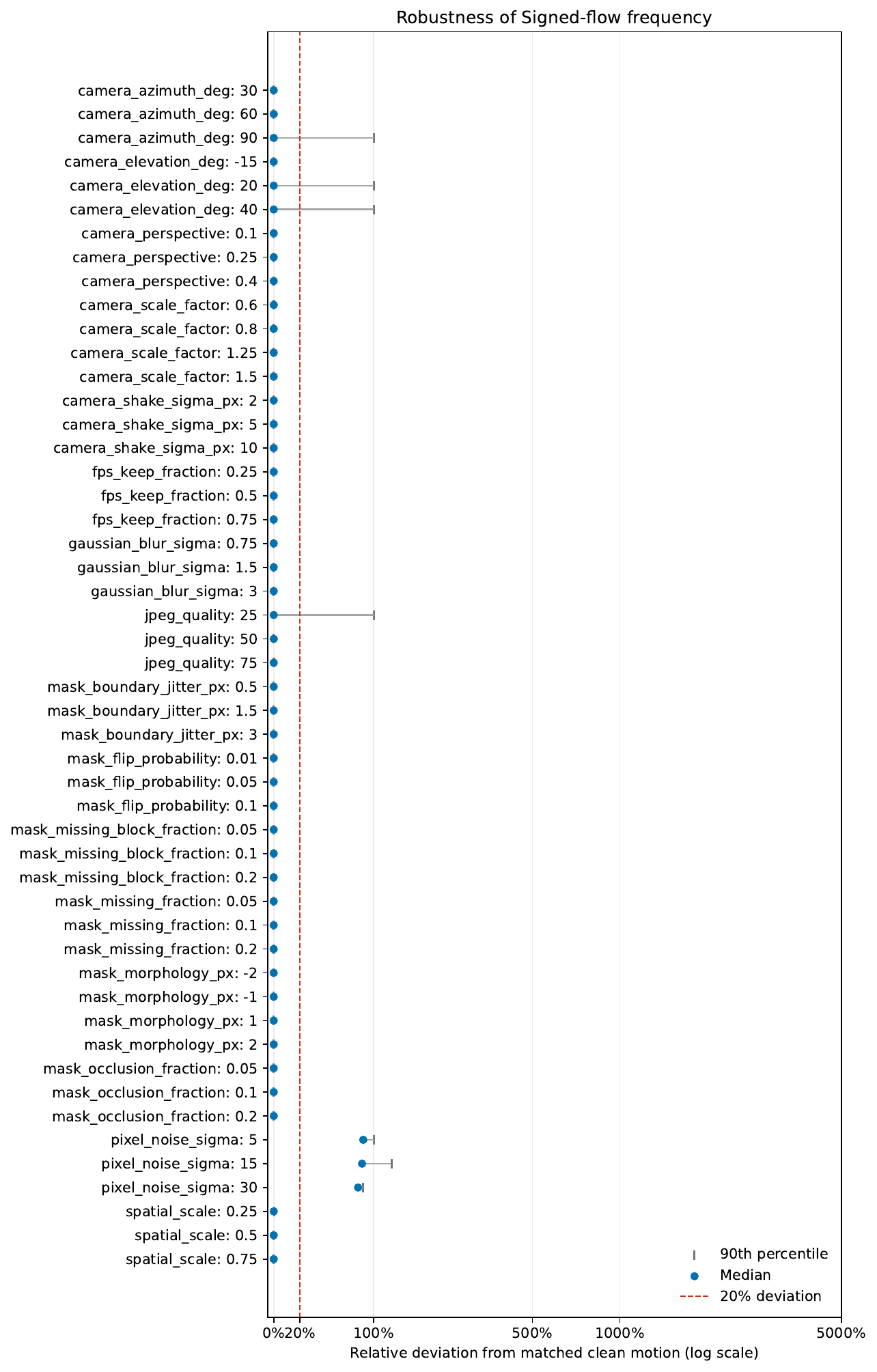}
        \caption{Estimated frequency.}
        \label{fig:freq_robust}
    \end{subfigure}
    \caption[]{\textbf{Sensitivity to observation perturbations, continued.} Estimated frequency under the tested observation perturbations, compared with matched unperturbed counterparts.}
\end{figure*}

\clearpage

\begin{figure*}[p]
    \ContinuedFloat
    \centering
    \captionsetup[subfigure]{font=small,skip=5pt}
    \begin{subfigure}{0.7\linewidth}
        \centering
        \includegraphics[width=\linewidth]{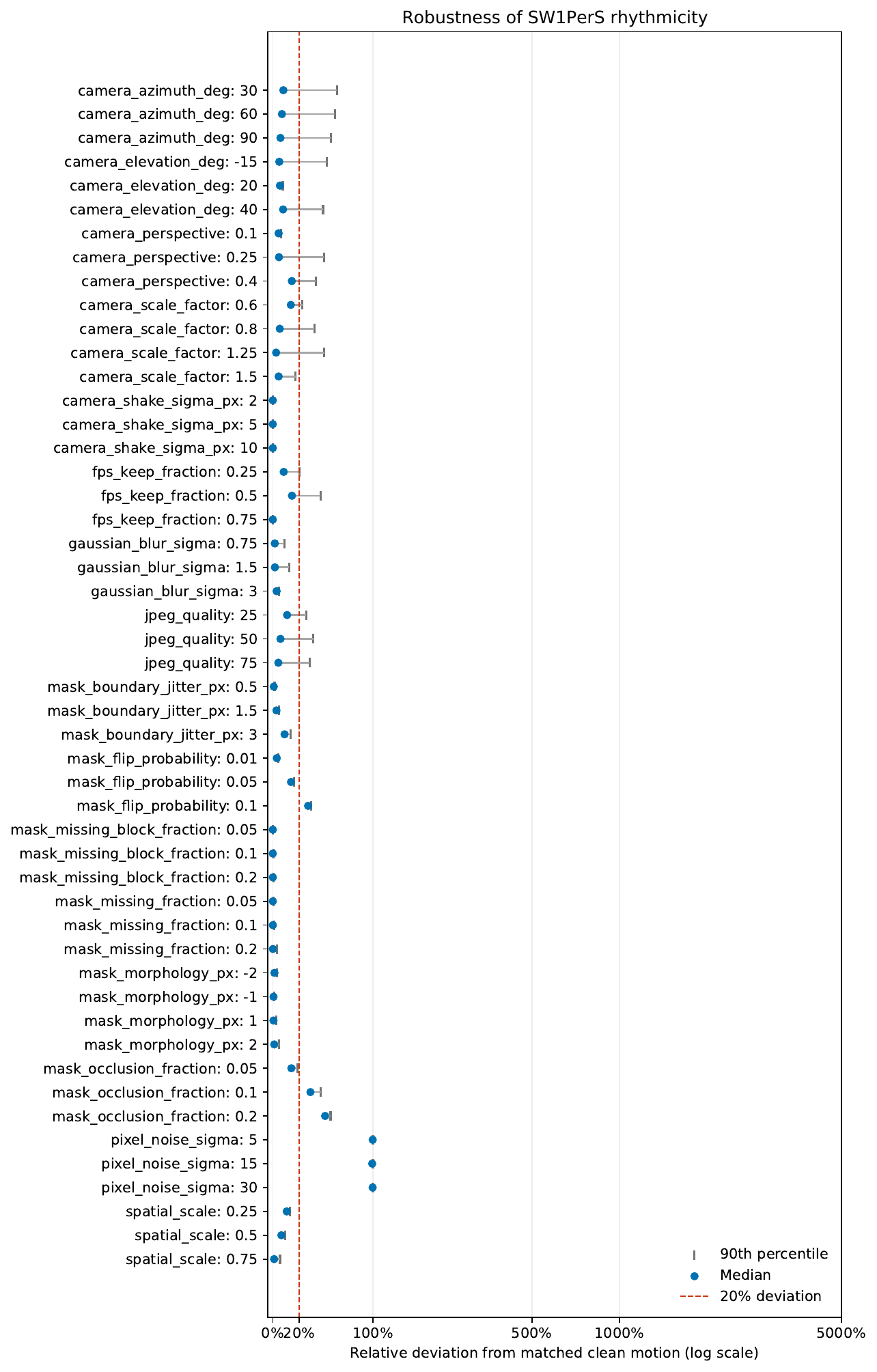}
        \caption{Rhythmicity score, $PS$.}
        \label{fig:periodicity_robust}
    \end{subfigure}
    \caption[]{\textbf{Sensitivity to observation perturbations, continued.} Rhythmicity under the tested observation perturbations, compared with matched unperturbed counterparts.}
\end{figure*}

\clearpage

\section{Sparse Landmark Observation and Identifiability}
\label{app:sparse-observation}
We consider what a fixed set of landmark velocities reveals about a dense field. At one frame, the field has $P$ spatial samples with two components each, so $F\in\mathbb R^{2P}$. Fix $J<P$ landmark locations and sampling weights. Point sampling or fixed local averaging defines a linear map $S:\mathbb R^{2P}\to\mathbb R^{2J}$. This observation model concerns sampled velocities; anatomical pose estimation from RGB images can use additional information.

\begin{proposition}[Sparse observations and recoverable functionals]
    \label{prop:sparse-observation}
    The operator satisfies:
    \begin{equation}
        \dim\ker S=2P-\operatorname{rank}S\ge2P-2J>0.
    \end{equation}
    Moreover, $SF=SG$ if and only if $G-F\in\ker S$. For any admissible set $\mathcal A\subseteq\mathbb R^{2P}$ and functional $Q:\mathcal A\to\mathbb R$, there exists $q:S(\mathcal A)\to\mathbb R$ with $Q=q\circ S$ on $\mathcal A$ if and only if:
    \begin{equation}
        SF=SG\ \Longrightarrow\ Q(F)=Q(G),\quad F,G\in\mathcal A.
    \end{equation}
\end{proposition}
\begin{proof}
    Rank--nullity gives the equality, and the codomain dimension bounds $\operatorname{rank}S$ by $2J$. Since the map is linear, $SF=SG\iff S(G-F)=0$.

    If $Q=q\circ S$, equal observations must give equal functional values. For the converse, assume the stated implication. Given $v\in S(\mathcal A)$, choose any $F\in\mathcal A$ with $SF=v$ and set $q(v)$ to $Q(F)$. Such an $F$ exists because the chosen value lies in the image. If a second field $G$ produces the same $v$, the assumption gives $Q(F)=Q(G)$. The choice of field therefore makes no difference, and $q$ is well-defined with $q(SF)=Q(F)$ on $\mathcal A$.
\end{proof}

\paragraph{Interpretation.}
Sparse observations leave some dense fields indistinguishable. They still determine a particular measurement if that measurement takes the same value for all admissible fields with the same observations. This qualification matters: the restriction of $S$ to a physical class $\mathcal A$ may be injective even when the ambient kernel is nontrivial. To demonstrate ambiguity in a measurement, we need admissible $F,G$ with identical observations and different measurement values.

\paragraph{Sequences.}
The argument also applies to a finite sequence of $K$ frames. For fixed operators $S_t$, define $\mathcal S(F_1,\ldots,F_K)=(S_1F_1,\ldots,S_KF_K)$. This is a block-diagonal linear map on the product space, so the same factorization criterion applies to sequence measurements. With the discrete integration rule $q_{t+1}=q_t+\Delta t_tV_t$, equal sampled velocities, time steps, and initial positions produce equal sampled position trajectories by induction. Initial positions and anatomical identities must still be supplied separately.

RGB pose estimators can use information beyond optical flow, so this proposition does not rank their accuracy. It also provides neither a learned pose decoder nor a guarantee that flow summaries suffice for classification or clinical interpretation. The factorization statement establishes the existence of a function on the observations; continuity and stability require further assumptions.

\section{Measurement Operators and Their Properties}
\label{app:operators}
Here we establish properties of the measurements after their input observations have been constructed. Errors introduced by tracking, stabilization, flow estimation, or changing valid support must be considered separately. Vector norms are Euclidean unless marked $L^2$, and each discrete pixel has unit area.

\subsection{Relative intensity}
Fix the analyzed frames. At frame $t$, let $D_t$ be the nonempty retained support, $K_t$ the nonempty reference core, and $F_t(p)\in\mathbb R^2$ the flow in fixed normalized units. Define:
\begin{align}
    u_t&=\big(\operatorname{med}_{K_t}F_{t,1},\operatorname{med}_{K_t}F_{t,2}\big),\\
    q_t(F)&=Q_{.75}\{\|F_t(p)-u_t\|:p\in D_t\},\\
    I(F)&=\operatorname{med}_t q_t(F).
\end{align}
Both inputs use the same empirical median and quantile conventions.

\begin{proposition}[Relative-intensity properties]
    \label{prop:intensity-properties}
    Keep the frame indices, supports, cores, and normalization fixed. For any spatially constant vector $c_t$ in each frame and scalar $a>0$,
    \begin{equation}
        I(F+c)=I(F),\qquad I(aF)=aI(F).
    \end{equation}
    If, on $D_t\cup K_t$ for every frame,
    \begin{equation}
        \sup_{t,p}\|F_t(p)-F'_t(p)\|\leq\epsilon,
    \end{equation}
    then
    \begin{equation}
        |I(F)-I(F')|\leq(1+\sqrt2)\epsilon.
    \end{equation}
\end{proposition}
\begin{proof}
    Adding $c_t$ shifts the component-wise core median by the same amount, leaving $F_t-u_t$ unchanged. Multiplication by $a>0$ scales the core median, residual norms, spatial quantile, and temporal median by $a$. These observations prove the two equalities.

    For the perturbation bound, consider paired real samples with $|x_j-y_j|\leq r$. Their empirical quantiles differ by at most $r$: the entry-wise inequalities $x_j-r\leq y_j\leq x_j+r$, together with monotonicity and translation equivariance of quantiles, give this bound.

    Each coordinate of $F_t-F'_t$ is bounded in absolute value by $\epsilon$. Applying the quantile bound to the two core medians gives $\|u_t-u'_t\|\leq\sqrt2\epsilon$. The triangle and reverse triangle inequalities then give, at every retained point,
    \begin{equation}
        \begin{aligned}
            &\big|\|F_t(p)-u_t\|-\|F'_t(p)-u'_t\|\big|\\
            &\quad\leq\|F_t(p)-F'_t(p)\|+\|u_t-u'_t\|\\
            &\quad\leq(1+\sqrt2)\epsilon.
        \end{aligned}
    \end{equation}
    Applying the quantile bound to $q_t$ and then to the temporal median completes the proof.
\end{proof}

\paragraph{Interpretation.}
Subtracting the reference removes a spatially constant additive flow, and the bound controls the effect of uniformly small vector errors when support is fixed. Changing support or data-dependent preprocessing falls outside these assumptions. The component-wise median also depends on the coordinate axes, so this result does not imply invariance to arbitrary rotations or camera views.

\subsection{Silhouette extent}
For binary masks $M_1,\ldots,M_N$ on the same finite crop, write $\bar M=N^{-1}\sum_tM_t$ and, for a positive area normalizer $S$,
\begin{equation}
    E_S(M)^2=\frac{1}{NS}\sum_t\|M_t-\bar M\|^2.
\end{equation}
In the code, $S$ is the median area of the retained nonempty masks, with a one-pixel floor. Invalid and empty masks are omitted. We compare the same retained frames below; the final bound also allows the area normalizers to differ.

\begin{proposition}[Extent as pairwise silhouette change]
    \label{prop:extent-properties}
    For these masks,
    \begin{equation}
        E_S(M)^2=\frac{1}{2N^2S}\sum_{t,s}|M_t\triangle M_s|.
    \end{equation}
    If $r_t=|M_t\triangle M'_t|$ and the normalizer is fixed, then:
    \begin{equation}
        |E_S(M)-E_S(M')|\leq\sqrt{\frac{\sum_t r_t}{NS}}.
    \end{equation}
    For separately estimated positive normalizers $S,S'$, one has:
    \begin{equation}
        |E_S(M)-E_{S'}(M')|\leq\sqrt{\frac{\sum_t r_t}{NS}} + E_{S'}(M')\left|\sqrt{S'/S}-1\right|.
    \end{equation}
\end{proposition}
\begin{proof}
    Expanding squared distances yields:
    \begin{equation}
        \begin{aligned}
            \sum_{t,s}\|M_t-M_s\|^2
            &=2N\sum_t\|M_t\|^2-2\Big\|\sum_tM_t\Big\|^2\\
            &=2N\sum_t\|M_t-\bar M\|^2.
        \end{aligned}
    \end{equation}
    For binary masks, squared distance equals the number of differing pixels. Substituting this fact proves the identity.

    Stack the masks as $m=(M_1,\ldots,M_N)$, and let $H$ subtract their temporal mean from each block. The map $H$ is an orthogonal projection, so $\|H(m-m')\|\leq\|m-m'\|$. Since $\|m-m'\|^2=\sum_t r_t$, we obtain:
    \begin{equation}
        \begin{aligned}
            |E_S(M)-E_S(M')|
            &=\frac{|\|Hm\|-\|Hm'\||}{\sqrt{NS}}\\
            &\leq\frac{\|m-m'\|}{\sqrt{NS}}.
        \end{aligned}
    \end{equation}
    This proves the bound for a fixed normalizer. For different normalizers, insert $E_S(M')$ in the triangle inequality and use $E_S(M')=\sqrt{S'/S}E_{S'}(M')$.
\end{proof}

\paragraph{Interpretation.}
Extent summarizes projected silhouette variation. The bounds control absolute changes; relative changes can still be large when clean extent is close to zero. They also do not convert silhouette variation into physical three-dimensional excursion.

\section{Oscillator Models and Supplementary Examples}
\label{app:oscillator-examples}
We use two simple models to examine what the measurements capture. The first shows why intensity, extent, and frequency carry complementary information. The second tests temporal organization using Eulerian field inner products. Both models make explicit assumptions about movement and observation; neither establishes that four measurements identify arbitrary motion or its underlying oscillator.

\subsection{Example 1: three-factor identification}
Let $R_0\subset\mathbb R^2$ be a stationary region of area $B>0$. A rectangle $R_\theta$ of horizontal width $w>0$ and height $\ell>0$ moves by translation $X(\theta)=A\sin\theta$, with its entire sweep disjoint from $R_0$. Let:
\begin{equation}
    \theta(t)=2\pi ft+\phi,\qquad f>0,\quad 0<2A<w.
\end{equation}
We observe the entire swept mask, with exact masks and instantaneous velocities. Spatial sampling is uniform in area, and phase $\Theta$ is uniform on $[0,2\pi)$. The reference core lies in $R_0$, and the reference height $h>0$ is fixed and known. Set:
\begin{equation}
    S=B+w\ell,\quad \frac{w\ell}{S}>\frac14,\quad c=\frac{\sqrt2\pi}{h},\quad k=\frac8{\pi^2}.
\end{equation}
Use the population spatial quantile $Q_q(Y)=\inf\{y:\Pr(Y\leq y)\ge q\}$. Define $I$ as the phase median of the normalized spatial upper quartile of speed, and let:
\begin{equation}
    E^2=\frac{\mathbb E\|M_\Theta-\mathbb E M_\Theta\|_{L^2}^2}{S}, \quad M_\Theta=\mathbf1_{R_0\cup R_\Theta}.
\end{equation}

\begin{proposition}[Three-factor population identification]
    \label{prop:oscillator-identification}
    Under these assumptions, if frequency is recovered exactly as $F=f$,
    \begin{equation}
        I=cAf,\qquad E^2=\frac{kA\ell}{B+w\ell},\qquad F=f.
        \label{eq:supp-oscillator-map}
    \end{equation}
    For known $B,w,h$, the measurement triple uniquely determines $(A,f,\ell)$ on this parameter domain, with inverse:
    \begin{equation}
        f=F,\quad A=\frac{I}{cF},\quad \ell=\frac{BE^2}{kA-wE^2}.
        \label{eq:supp-oscillator-inverse}
    \end{equation}
\end{proposition}
\begin{proof}
    \emph{Intensity.} At each phase, the moving points share speed $2\pi fA|\cos\theta|$, while stationary points have zero speed. Because the stationary fraction is less than $3/4$, the spatial upper quartile equals the moving speed, including at phases where it is zero. The stationary reference core contributes nothing to the subtraction. For uniform phase, $\operatorname{med}|\cos\Theta|=1/\sqrt2$, since $\Pr(|\cos\Theta|\leq1/\sqrt2)=1/2$. Hence $I=cAf$.

    \emph{Extent.} Take independent uniform phases $\Theta,\Theta'$. Expanding the squared norm gives the population variance identity
    \begin{equation}
        E^2=\frac{1}{2S}\mathbb E \|M_\Theta-M_{\Theta'}\|_{L^2}^2.
    \end{equation}
    The stationary region cancels from the difference. Since $|X(\Theta)-X(\Theta')|\le2A<w$, the symmetric difference of the rectangles has area $2\ell|X(\Theta)-X(\Theta')|$. Thus
    \begin{equation}
        E^2=\frac{\ell}{S}\mathbb E|X-X'|, \quad X=A\sin\Theta.
    \end{equation}
    On $[-A,A]$, the displacement has distribution function $G(x)=1/2+\arcsin(x/A)/\pi$. For bounded independent copies,
    \begin{equation}
        \mathbb E|X-X'|=2\int_{-A}^{A}G(x)(1-G(x))\,dx.
    \end{equation}
    To see this identity, write $|X-X'|$ as the integral over $x$ of the indicator that $x$ lies between $X$ and $X'$. Its expectation is $2G(x)(1-G(x))$. Substituting $x=A\sin u$ gives
    \begin{equation}
        \begin{aligned}
            \mathbb E|X-X'|&=2A\int_{-\pi/2}^{\pi/2}\left(\frac14-\frac{u^2}{\pi^2}\right)\cos u\,du\\
            &=2A\left[\frac12-\frac{\pi^2/2-4}{\pi^2}\right] =\frac{8A}{\pi^2}.
        \end{aligned}
    \end{equation}
    Two integrations by parts give $\int_{-\pi/2}^{\pi/2}u^2\cos u\,du=\pi^2/2-4$, completing the extent calculation.

    \emph{Uniqueness.} First recover $f$ from $F$, then $A$ from $I$. Rearranging the extent expression gives Eq.~\eqref{eq:supp-oscillator-inverse}. For measurements generated by the model,
    \begin{equation}
        kA-wE^2=\frac{kAB}{B+w\ell}>0,
    \end{equation}
    The denominator is therefore positive, and the inverse is well defined. Any two admissible triples with the same measurements must have the same parameters.
\end{proof}

\noindent
\begin{minipage}{\columnwidth}
    \begin{proposition}[Non-redundancy within the model]
        \label{prop:oscillator-nonredundancy}
        Fix known $B,w,h>0$ and the open admissible parameter set of Proposition~\ref{prop:oscillator-identification}. Through every admissible $(A,f,\ell)$, each pair selected from $(I,E,F)$ has a non-constant local curve of parameter triples with identical paired measurements. Consequently, no such pair identifies all three parameters on this set.
    \end{proposition}
\end{minipage}
\begin{proof}
    Write $i=I>0$ and $e=E^2>0$ at the chosen triple. For the pair $(I,F)$, hold $A,f$ fixed and vary $\ell$. For $(E,F)$, hold $f$ fixed and vary $u=\ell$, setting
    \begin{equation}
        A(u)=\frac{e(B+wu)}{ku}.
    \end{equation}
    This preserves $E^2=e$ and $F=f$ and passes through the original triple. For $(I,E)$, vary $a=A$ and set
    \begin{equation}
        f(a)=\frac{i}{ca},\qquad \ell(a)=\frac{Be}{ka-we}.
    \end{equation}
    The identification proof gives a positive denominator at the original amplitude, so the curve is continuous in a neighborhood of that amplitude. Substitution yields $I=i$ and $E^2=e$. Each curve changes its free parameter and is therefore non-constant. Since all constraints are strict, continuity keeps the curve admissible on a sufficiently small interval. The observation domain must continue to contain the full sweep, and the reference core must remain valid throughout that interval.
\end{proof}

Each coordinate is necessary to identify the target $(A,f,\ell)$ from this measurement triple. Removing any one leaves a local ambiguity.

\paragraph{What the result explains.}
Intensity combines amplitude and frequency, so either can compensate for a change in the other. Knowing frequency separates them, and extent then determines the moving-region geometry. Full-cycle population averaging discards initial phase. The explicit inverse establishes global uniqueness within this family, although recovery can be poorly conditioned: a small $kA-wE^2$ amplifies errors in extent. The Jacobian of $(I,E^2,F)$ with respect to $(A,f,\ell)$ has determinant:
\begin{equation}
    -\frac{ckAfB}{(B+w\ell)^2}\ne0,
\end{equation}
which also establishes local differential invertibility. The explicit inverse above is what establishes global uniqueness. Both results assume accurate frequency and known reference geometry. At $B=0$, extent no longer depends on $\ell$; when the moving fraction is below $1/4$, intensity is zero. Without moving area or excursion, the observations contain no motion frequency.

\begin{example}[Three-factor recovery]
    \label{ex:identification-video}
    Supplementary Video~1 uses $B=w=1$, $h=3$, and baseline $(A,f,\ell)=(.10,1,.80)$. Three six-second scenes separately change $A$ to $.18$, $f$ to $1.70$ Hz, and $\ell$ to $1.40$. Each scene displays the exact population measurements and the parameters recovered from them. The baseline gives $(I,E,F)\approx(.148096,.189803,1)$. These displayed values come from the population model, which assumes complete-cycle averaging and $F=f$; they are not estimated from the six-second excerpts.
\end{example}

\subsection{Example 2: matched Eulerian temporal organization}
\label{app:matched-eulerian}
Now introduce phase modulation in the same rectangle model:
\begin{align}
    \theta(t)&=2\pi ft+\phi+\varepsilon\sin(2\pi\nu t+\psi),\\
    X_t&=A\sin\theta(t),\\
    V_t&=A\big[2\pi f+2\pi\nu\varepsilon \cos(2\pi\nu t+\psi)\big]\cos\theta(t),
\end{align}
where $\nu=.37\sqrt2$ Hz. When $\varepsilon>0$, the phase and speed distributions over the record change. We therefore match the finite-record measurements directly, rather than applying the pure-sinusoid population inverse unchanged.

We set $B=w=1$ and $h=3$, sampling 720 times at 60 Hz over 12 s. Eight phase seeds and $\varepsilon\in\{0,.15,.3,.6\}$ give 32 conditions. At each candidate frequency, we rescale amplitude to match the clean finite-record intensity.

The rectangle model gives exact finite-record measurements and an Eulerian field Gram matrix:
\begin{align}
    I&=\operatorname{med}_t|V_t|/h,\\
    E^2&=\frac{\ell}{B+w\ell}\operatorname{mean}_{t,s}|X_t-X_s|,\\
    G_{ts}&=\frac{\ell(w-|X_t-X_s|)V_tV_s}{h^2}.
\end{align}
Because the displacement range is below $w$, the overlap area of two moving rectangles is $\ell(w-|X_t-X_s|)$. The Gram matrix therefore accounts for spatial overlap when taking Eulerian field inner products. It is not the Gram matrix of one fixed material vector scaled by a sinusoid. Individual entries may be negative while the full matrix remains positive semi-definite.

For each candidate $f$, we multiply amplitude by the ratio of baseline to candidate intensity. With $\mu=\operatorname{mean}_{t,s}|X_t-X_s|$ and baseline squared extent $e$, we set $\ell=Be/(\mu-we)$. A bracketed root solve on $[.9,1.1]$ Hz then matches the frequency estimated from the mean-distance lag profile, with tolerance $10^{-9}$ Hz. Persistence does not enter this matching objective. Each of the eight seeds determines initial and modulation phases using random seed $2026101500+\mathrm{seed}$. We retain every requested modulation level.

We compute PS from the exact field Gram matrix and estimated period, using $d=23$, 300 windows, within-window centering, sphere normalization, and Vietoris--Rips $H_1$ over $\mathbb F_{47}$. This calculation bypasses image estimation, crop resampling, valid-pixel weighting, and winsorization. Compensation controls keep the adjusted amplitude, rectangle height, carrier, and initial phase while removing modulation. We score them at both the matched reference period and their own estimated period.

As modulation increases, mean $PS$ falls through $.949,.788,.672,.514$, while mean cycle-duration CV rises through approximately $0,.034,.068,.135$, respectively. Every irregular condition scores below its clean baseline, and scores decrease at every level for every seed. Intensity and extent remain matched to relative error below $3\times10^{-16}$, and estimated frequency to below $2\times10^{-10}$. Removing modulation while keeping the adjusted geometry and carrier raises all 24 compensation scores above their irregular counterparts ($PS$ range $.931$--$.949$).

\begin{example}[Matched organization]
    \label{ex:matched-video}
    Supplementary Video~2 compares phase seed 0 at $\varepsilon=0$ and $.6$. Both have $(I,E,F)\approx(.148064,.189791,.999821)$, but their $PS$ values are $.949197$ and $.519563$, respectively. Their amplitudes and moving heights differ because matching holds the measurements fixed while allowing generating parameters to change. The labels show the archived measurements from the 60-Hz records, and playback is 30 fps. This pair illustrates one seed of the full experiment.
\end{example}

\subsection{Transfer through rendered motion}
We also test the image-based pipeline on body rocking, analyzing 12 s at 20 fps after a two-second warm-up. One family introduces stochastic timing modulation with correlation scale .25 s; the other adds a $\sqrt2$-Hz rhythm. Strengths $0,.25,.5,.75$ and three seeds give 24 design records, each evaluated at estimated and reference periods for 48 evaluations. Clean records are shared between families, and the quasi-periodic seed copies are numerically identical. There are therefore 13 distinct rendered bundles, not 24 independent physical replications.

Under stochastic disruption, median scores using estimated periods are $.884,.702,.520,.383$; the reference-period values are $.883,.688,.532,.347$. For the competing rhythm, estimated-period medians are $.884,.653,.423,.393$, compared with $.883,.653,.423,.156$ at the reference period. The strongest mixture changes which rhythm is selected. These rendered controls show how the score responds through the image pipeline, while allowing intensity, extent, and frequency to vary. The evidence for additional information at matched values comes from the exact Eulerian controls.

\section{Real Profiles, Matched Bouts, and Pose Convergence}
\label{app:real-profiles}

We analyze 1,404 unique annotated bouts from six participants across 26 source files, keeping the expert boundaries fixed. Duration is the difference between annotated start and end times. Within each recording, the preceding inter-bout interval runs from the previous annotated bout's offset to the current onset, regardless of label. First bouts have no preceding interval, and negative intervals are recorded as overlaps rather than converted to gaps. There are 1,378 available nonnegative intervals.

\begin{table*}[t]
    \small
    \centering
    \caption{\textbf{Real-video distributions.} Median [Q25,Q75] for each available measurement in the primary frequency analysis. Intensity is in approximate tracked body heights/s, frequency in Hz, and duration and IBI in seconds. Extent measures area-normalized silhouette RMS variation. Available bout counts differ across measurements.}
    \resizebox{\textwidth}{!}{
    \begin{tabular}{lcccccc}
        \toprule
        Behavior & Intensity & Extent & Frequency (Hz) & $PS$ & Duration (s) & IBI (s) \\
        \midrule
        Flap & 0.094 [0.062,0.135] & 0.319 [0.267,0.380] & 0.641 [0.363,1.119] & 0.032 [0.012,0.058] & 2.586 [1.663,4.381] & 3.511 [1.037,10.715] \\
        Flap--rock & 0.192 [0.123,0.270] & 0.482 [0.395,0.567] & 1.141 [0.849,1.682] & 0.131 [0.073,0.236] & 4.288 [2.939,6.457] & 2.501 [0.439,8.505] \\
        Rock & 0.173 [0.114,0.239] & 0.418 [0.349,0.533] & 0.763 [0.662,0.836] & 0.352 [0.187,0.477] & 6.241 [3.134,12.943] & 4.421 [1.563,13.984] \\
        \bottomrule
    \end{tabular}}
    \label{tab:real-profiles}
\end{table*}

\begin{table}[t]
    \small
    \centering
    \caption{\textbf{Measurement availability.} Available counts among all bouts with each label. Frequency uses the primary quality-control rule.}
    \begin{tabular}{lrrrrr}
        \toprule
        Label & Total & $I$ & $E$ & $\widehat f$ & $PS$ \\
        \midrule
        Flap & 616 & 616 & 616 & 199 & 218 \\
        Flap--rock & 188 & 186 & 188 & 76 & 111 \\
        Rock & 600 & 597 & 599 & 440 & 433 \\
        \bottomrule
    \end{tabular}
    \label{tab:availability}
\end{table}

\subsection{Within-recording matching}

We match bouts within participant, behavior, and recording, expressing each measurement as a percentile rank within that group. A pair must lie within six percentile points on the matching coordinate; we then check whether another coordinate differs by at least 25 percentile points. The primary analysis contains 4,909 intensity-matched pair records, 5,356 extent-matched, 1,488 frequency-matched, and 1,493 organization-matched records. Of these, $52.1\%$, $53.8\%$, $67.3\%$, and $67.5\%$, respectively, meet the contrast criterion on another coordinate. Pairs can share bouts and can appear under more than one matching coordinate, so these proportions describe overlapping examples rather than independent events or error-controlled tests.

The main figure shows two P001 rocking bouts from one recording. The accompanying table retains their labels, selection rule, episode identifiers, matching ranks, and measurements. They are matched on intensity and have similar frequency, while extent and $PS$ differ. This selected pair illustrates what the profile adds to the behavior label. It does not estimate the prevalence of that pattern, and observation error and matching conditions still affect the measurements.

\subsection{Repeated-session summaries}

For P001 and P004, we summarize each session using behavior-specific medians and bout-level interquartile ranges. A session summary for a measurement requires at least three available bouts. Recordings made on the same day keep separate run identifiers. Setting, viewpoint, elapsed time, and participant state can vary together, so the plots describe the recorded sessions without isolating intervention or developmental effects.

\subsection{Matched-support pose convergence}
\label{app:pose-convergence}

For each pose--flow comparison, we use only frame pairs valid for both measurements. We compare total pose motion with total flow. Articulated pose motion subtracts the torso reference and is compared with torso-corrected flow, while torso translation is evaluated separately. Mask height supplies a common image scale, and confidence filtering and pose-support requirements determine which pairs are retained. Matching the frame pairs makes their observational support comparable, although neither source supplies ground-truth velocity.

\begin{table*}[t]
    \small
    \centering
    \caption{\textbf{Pose--flow agreement.} Pooled Spearman correlations and correlations of ranks within participant and behavior. Intervals are participant-clustered 95\% bootstrap intervals from 2,000 successful replicates.}
    \begin{tabular}{lrrr}
        \toprule
        Comparison & Bouts & Pooled $\rho$ [95\% CI] & Within-stratum rank association [95\% CI] \\
        \midrule
        Total motion & 1379 & 0.817 [0.680,0.907] & 0.764 [0.715,0.820] \\
        Articulated / residual & 1376 & 0.695 [0.548,0.777] & 0.657 [0.578,0.732] \\
        Torso translation & 1376 & 0.734 [0.544,0.876] & 0.628 [0.554,0.706] \\
        \bottomrule
    \end{tabular}
    \label{tab:pose-convergence}
\end{table*}

For total motion, pooled correlation is $.817$ with interval $[.680,.907]$, and correlation after ranking within participant and behavior is $.764$ with interval $[.715,.820]$. The corresponding values are $.695$ and $.657$ for articulated pose and residual flow, and $.734$ and $.628$ for torso translation. The bootstrap still has only six participant clusters despite the many bouts. Both sources also share video, mask support, and height normalization. These associations therefore quantify agreement, without establishing absolute accuracy or validating every PRISM measurement.

\section{Pose Observation Audits}
\label{app:pose-audit}

We examine three aspects of pose observation: confidence and availability, temporal continuity, and sensitivity to pre-processing. AQSM MediaPipe and SAM-crop pose cover the same 1,404 bouts. ASDMotion contributes 36,930 external units, which are analyzed separately because their sampling design differs from the bout cohort. We report confidence thresholds $.3,.5,.7$ within each system; the scores are not calibrated across systems.

\begin{table*}[t]
    \small
    \centering
    \caption{\textbf{Pose coverage after confidence filtering.} Median fraction of retained landmark observations per unit at threshold $c$. The two AQSM sources cover the same cohort, while ASDMotion is external. Thresholds use each system's own confidence scale, and coverage does not measure accuracy.}
    \begin{tabular}{lrrrr}
        \toprule
        Source & Units & $c=.3$ & $c=.5$ & $c=.7$ \\
        \midrule
        MediaPipe, same cohort & 1,404 & 76.5\% & 73.8\% & 64.7\% \\
        SAM-crop pose, same cohort & 1,404 & 69.9\% & 64.2\% & 58.3\% \\
        ASDMotion, external & 36,930 & 57.9\% & 51.4\% & 39.8\% \\
        \bottomrule
    \end{tabular}
    \label{tab:pose-coverage}
\end{table*}

At threshold .5, median landmark coverage is $.738$ for MediaPipe, $.642$ for SAM-crop pose, and $.514$ for ASDMotion. Motion measurements need consecutive observations: median valid frame-pair coverage is $.733$, $.608$, and $.499$, respectively. We also retain wrist coverage and joint-level denominators, since overall coverage can hide missing observations in a particular region.

We flag gaps and abrupt displacements in the original trajectories before interpolation or smoothing. A flag marks a sequence for inspection: rapid true movement, camera effects, occlusion, or coordinate conventions can produce unusual displacements and estimation errors. The pose-sequence illustration provides visual context without labeling every flagged point as incorrect.

\begin{figure*}[htb]
    \centering
    \includegraphics[width=.9\textwidth]{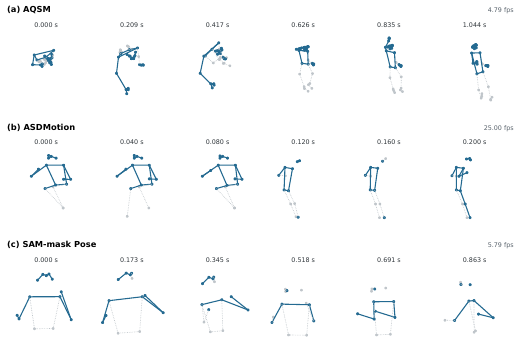}
    \caption{\textbf{Illustrative pose sequences.} Examples give context for the coverage and continuity audits. Abrupt changes are candidates for inspection, not verified pose errors; quantitative counts are reported in the audit tables.}
    \label{fig:pose-sequences}
\end{figure*}

We vary confidence filtering, short-gap interpolation, and temporal smoothing before computing total and articulated motion. This demonstrates how processing choices affect the summaries; it does not establish which choice recovers true movement. The sensitivity table retains the settings for each unit, selected-unit identifiers, valid motion-pair coverage, baseline values at the same threshold, and changes from those baselines. Units without a usable normalization scale are listed separately.

The processing table contains 69,111 records from 1,404 MediaPipe units, 1,389 SAM-crop units, and 2,093 external units. Confidence thresholds are $.3,.5,.7$, and the maximum interpolation gap is 0 or .2 s. We compare unsmoothed trajectories with smoothing at the scales recorded for each source. AQSM uses .1 and .2 s settings; the external records also include frame-rate-specific .0333/.0667 and .04/.08 s settings. Summaries report the number of valid units at each setting. Multiple settings on the same unit do not add independent observations.

\section{Supplementary Landmark Multi-view Study}
\label{app:multiview}

We examine how landmark-derived measurements change with observation geometry, separately from the flow and mask measurements PRISM uses. The 1,380 trajectory files are evaluated using exact 3D observations, favorable and unfavorable single views chosen from the generating geometry, random views, and calibrated multi-view reconstruction. Calibration with held-out observation seeds tests recovery within this simulator design. The favorable view uses knowledge of the generating geometry and has not been validated as an automatic selection procedure.

\begin{table}[t]
    \small
    \centering
    \caption{\textbf{Landmark recovery across observation conditions.} Pooled median calibrated relative error, expressed as a percentage. This experiment evaluates landmark measurements separately from the main PRISM profiles.}
    \begin{tabular}{lrrr}
        \toprule
        Observation & Intensity & Extent & Frequency \\
        \midrule
        Calibrated multiview & 12.80 & 3.61 & 0.20 \\
        Favorable single view & 14.87 & 1.11 & 0.24 \\
        Random single view & 17.76 & 6.20 & 0.20 \\
        Unfavorable single view & 21.35 & 4.32 & 0.45 \\
        \bottomrule
    \end{tabular}
    \label{tab:multiview}
\end{table}

In the pooled paired summaries, calibrated multi-view reconstruction modestly improves intensity and frequency over the favorable single view, while the favorable view gives better extent estimates overall. Arm flapping shows a particularly clear difference: multi-view extent has a median relative error of $21.43\%$ and a recovery correlation of $.375$, compared with $1.04\%$ and $.927$ for the favorable view. Reconstruction therefore does not improve every coordinate in this design.

Rhythmicity recovery also varies with the observation. For arm flapping, the median within-condition recovery correlation is $.2$ with calibrated multi-view and $.8$ with the favorable view; the recorded paired organization difference is zero. Calibration, movement geometry, and visible motion can all affect this comparison. Multi-view acquisition may help under occlusion or foreshortening, but its benefit requires evaluation under those conditions.

\section{Profile-Conditioned Movement and Sample Size}
\label{app:exemplars}

We generate illustrative movements from a target profile and a known behavior family. The profile leaves out anatomy, initial phase, appearance, and many differences between motion fields. By convention, choosing a canonical skeleton, viewpoint, and motion family supplies that missing information. We then measure the generated movement with PRISM to see how closely it matches the target. This evaluates the example's numerical agreement; anatomical reconstruction, clinical comprehension, and privacy protection remain separate questions.

\subsection{Nested design and held-out targets}

The original library contains 48 clips per behavior, with 18 test clips each for body rocking, arm flapping, and hand flapping. We keep these 144 library clips and 54 test clips, then expand the library to 576 and 1,440 clips using nested scrambled Sobol samples of frequency input, amplitude scale, and disruption. Samples remain within the original bounds $(.75,.6,0)$ to $(2,1.4,.75)$. Added training conditions exclude the exact test values on each parameter axis and all test-generating seeds. New seeds produce the added simulator conditions. 

The shared input pool contains 1,494 conditions. Each smaller library is a subset of the largest, and all models are evaluated on the same 54 target profiles. Because the test parameters lie within the simulator's represented range, the study measures interpolation within the known motion families. We already inspected the fixed targets during earlier development, so they do not constitute an external validation set. Architecture and optimizer settings remain fixed across library sizes.

\subsection{Fixed neural model and evaluation}

The network takes four measurements and three behavior indicators as input. Two hidden layers, each with 16 tanh units, feed three linear outputs for simulator frequency, amplitude scale, and disruption, giving 451 trainable scalars. Input and output standard scalers are fitted only on the corresponding training library. We train with L-BFGS, regularization $\alpha=.01$, and a limit of 3,000 iterations. Before rendering, we clip predictions to the original parameter bounds and report how often clipping is needed.

Each clip has eight analyzed seconds at 20 fps following a two-second warm-up. We render from a canonical $45^\circ$ camera and measure at $256\times144$ pixels using the simulator silhouette. The original library crosses frequency inputs $.75,1,1.5,2$, amplitude multipliers $.6,1,1.4$, and disruptions $0,.25,.5,.75$. Tests use $.875,1.25,1.75$, $.8,1.2$, and $.125,.375,.625$, respectively. Disruption changes frequency along a smooth stochastic trajectory, so the input frequency need not equal the recurrence rate measured from the resulting clip.

We render every neural prediction with a new seed and measure the output using the same fixed pipeline. We provide all 1,494 input profiles and all 162 outputs from the three models. Validation checks the design, excluded test-axis values, separation of images and seeds, training IDs, network architecture, and scoring of the measured outputs.

For behavior $g$, let $s_{gk}$ be the IQR of coordinate $k$ in the \emph{original 144-library} profiles. We keep these scales fixed across library sizes and methods. The error between measured output $\widehat\Phi$ and target $\Phi$ is
\begin{equation}
    e(\widehat\Phi,\Phi)=\left[\frac14\sum_{k=1}^{4}\left(\frac{\widehat\Phi_k-\Phi_k}{s_{gk}}\right)^2\right]^{1/2}.
\end{equation}
Four-coordinate retrieval chooses the same-behavior library clip with the smallest profile error. Retrieval without $PS$ uses only the first three scaled coordinates to select a clip, then evaluates the full profile. The random baseline averages errors over 100 same-behavior draws per target with a fixed random state. We re-rank retrieval candidates at each size using the common scales. This IQR normalization defines the comparison metric and has no established clinical calibration.

\begin{table}[t]
    \small
    \centering
    \caption{\textbf{Neural sample-size comparison.} Architecture, target profiles, and reference scales are held fixed. Frequency MAE is measured in Hz.}
    \begin{tabular}{rrrrr}
        \toprule
        Training & Median error & $f$ MAE & $PS$ MAE & Clipped \\
        \midrule
        144 & 0.196 & 0.127 & 0.110 & 15 \\
        576 & 0.135 & 0.108 & 0.086 & 3 \\
        1440 & 0.087 & 0.114 & 0.070 & 0 \\
        \bottomrule
    \end{tabular}
    \label{tab:sample-size}
\end{table}

Expanding the library from 144 to 1,440 clips reduces median profile error from $.196409$ to $.086928$ and $PS$ MAE from $.109981$ to $.070286$. Clipping falls from 15 predictions to zero. Frequency MAE is lowest with 576 samples. At the descriptive .35 error cutoff, 45, 47, and 47 of the 54 targets fall within the cutoff, while 90th-percentile errors are $.429,.407,.361$. 

\begin{table}[t]
    \small
    \centering
    \caption{\textbf{Comparison with 1,440 library clips.} All methods use the same 54 targets and reference scales. Frequency MAE is in Hz; the last column counts profile errors at most .35.}
    \setlength{\tabcolsep}{5pt}
    \begin{tabular}{lrrrr}
        \toprule
        Method & Median error & $f$ MAE & $PS$ MAE & $\leq .35$ \\
        \midrule
        Neural & 0.087 & 0.114 & 0.070 & 47/54 \\
        Random mean & 0.598 & 0.483 & 0.182 & 0/54 \\
        Retrieval & 0.068 & 0.058 & 0.020 & 54/54 \\
        Without PS & 0.230 & 0.032 & 0.159 & 35/54 \\
        \bottomrule
    \end{tabular}
    \label{tab:exemplar-methods}
\end{table}

With the largest library, retrieval gives lower median profile error, frequency MAE, and $PS$ MAE than the neural decoder. The decoder has lower extent MAE ($.000219$ versus $.001518$), so the ranking differs by coordinate. Retrieval selects candidates using the same distance used to evaluate them, making this a test of matching quality rather than independent evidence for the measurements' construct validity. Movement properties omitted from the profile remain unresolved by either approach.

\section{VLM Experiment}

Forty clips crossed five behaviors with frequency (1 or 2 Hz), amplitude multiplier (.6 or 1.4), and duration (3 or 5 s). Grok 4.20, Grok 4.3, Grok 4.7, and 4-bit Qwen3-VL 2B received timestamped RGB frames at 10 fps with temperature zero and no examples. PRISM received the same frames plus masks and was also run at 30 fps.

For landmark $j$ relative to the root, reference intensity and movement were
\begin{align}
     v_t^*&=\frac1{20}\sum_{j=1}^{20}
     \frac{\|\mathbf q_{t+1,j}-\mathbf q_{t,j}\|_2}{H\Delta t_t},\\
     I^*&=\operatorname{median}_t v_t^*,
     &M^*&=\sum_t v_t^*\Delta t_t.
\end{align}
Active-cycle count accumulated oscillator phase over intervals exceeding 5\% of peak amplitude. PRISM proxies were $\widehat N=\widehat fT$ and $\widehat M=\widehat IT$. We presented twenty pairs in both orders. Missing or invalid answers failed; references within 10\% were ties.

\section{Supplementary Media and Interpretation}
\label{app:media}

Supplementary Video 1 varies amplitude, frequency, and moving-region height separately in the population oscillator model and displays its exact predicted measurements. Supplementary Video 2 animates one exact-field matched pair alongside its recorded measurements; the scores are not recomputed from the playback. Supplementary Video 3 shows the two fixed held-out rocking targets and their neural examples from the 1,440-library model, with measurements computed from the rendered outputs. All videos use canonical synthetic bodies or shapes and show no original participant appearance.

The first two videos illustrate the restricted observation models. The third shows one motion family compatible with the requested profile. The files and figure tables identify target values, population predictions, and achieved measurements separately. These illustrations do not evaluate anatomical reconstruction, clinical comprehension, or whether a profile preserves privacy.

%% file: main.bib
@String(CVPR= {IEEE Conf. Comput. Vis. Pattern Recog.})

@String(ICCV= {Int. Conf. Comput. Vis.})

@String(CVPR  = {CVPR})

@String(ICCV  = {ICCV})

@book{DSM5,
author={{American Psychiatric Association}},
title={Diagnostic and Statistical Manual of Mental Disorders},
edition={5},publisher={American Psychiatric Association},year={2013}}

@article{LaGrow1984,
author={LaGrow, Steven J. and Repp, Alan C.},
title={Stereotypic responding: A review of intervention research},
journal={American Journal of Mental Deficiency},year={1984},
volume={88},pages={595--609}}

@inproceedings{Albinali2009,
author={Albinali, Fahd and Goodwin, Matthew S. and Intille, Stephen S.},
title={Recognizing stereotypical motor movements in the laboratory and classroom: A case study with children on the autism spectrum},
booktitle={Proceedings of the 11th International Conference on Ubiquitous Computing},
year={2009},pages={71--80},doi={10.1145/1620545.1620555}}

@inproceedings{Goodwin2014,
author={Goodwin, Matthew S. and others},
title={Moving towards a real-time system for automatically recognizing stereotypical motor movements in individuals on the autism spectrum using wireless accelerometry},
booktitle={Proceedings of the 2014 ACM International Joint Conference on Pervasive and Ubiquitous Computing: Adjunct Publication},
year={2014},doi={10.1145/2632048.2632096}}

@article{Ravi2024SAM2,
author={Ravi, Nikhila and others},
title={{SAM 2}: Segment Anything in Images and Videos},
journal={arXiv preprint arXiv:2408.00714},year={2024}}

@article{Videnovic2025DAM4SAM,
author={Videnovic, Jovana and Kristan, Matej and Lukezic, Alan},
title={Distractor-Aware Memory-Based Visual Object Tracking},
journal={arXiv preprint arXiv:2509.13864},year={2025}}

@inproceedings{Kroeger2016DIS,
author={Kroeger, Till and Timofte, Radu and Dai, Dengxin and Van Gool, Luc},
title={Fast Optical Flow Using Dense Inverse Search},
booktitle={European Conference on Computer Vision},year={2016},
pages={471--488},doi={10.1007/978-3-319-46493-0_29}}

@article{PereaHarer2015,
author={Perea, Jose A. and Harer, John},
title={Sliding Windows and Persistence: An Application of Topological Methods to Signal Analysis},
journal={Foundations of Computational Mathematics},year={2015},
volume={15},pages={799--838},doi={10.1007/s10208-014-9206-z}}

@article{Tralie2018Ripser,
author={Tralie, Christopher and Saul, Nathaniel and Bar-On, Rann},
title={Ripser.py: A Lean Persistent Homology Library for Python},
journal={Journal of Open Source Software},year={2018},
volume={3},number={29},pages={925},doi={10.21105/joss.00925}}

@misc{AQSM2026,
author={MBaye, Austin A. and Perea, Jose A. and Tralie, Christopher J. and Goodwin, Matthew S.},
title={Automated Quantification of Stereotypical Motor Movements in Autism Using Persistent Homology},
year={2026}}

@article{wei2023vision,
title={Vision-based activity recognition in children with autism-related behaviors},author={Wei, Pengbo and Ahmedt-Aristizabal, David and Gammulle, Harshala and Denman, Simon and Armin, Mohammad Ali},journal={Heliyon},volume={9},number={6},year={2023}}

@article{negin2021vision,
title={Vision-assisted recognition of stereotype behaviors for early diagnosis of autism spectrum disorders},author={Negin, Farhood and Ozyer, Baris and Agahian, Saeid and Kacdioglu, Sibel and Ozyer, Gulsah Tumuklu},journal={Neurocomputing},volume={446},pages={145--155},year={2021}}

@article{Bauer2021Ripser,
author={Bauer, Ulrich},title={Ripser: efficient computation of {V}ietoris-{R}ips persistence barcodes},journal={Journal of Applied and Computational Topology},volume={5},year={2021},number={3},pages={391--423},doi={10.1007/s41468-021-00071-5}}

@InProceedings{Huang_2023_CVPR,
    author    = {Huang, Xiaofei and Luan, Lingfei and Hatamimajoumerd, Elaheh and Wan, Michael and Kakhaki, Pooria Daneshvar and Obeid, Rita and Ostadabbas, Sarah},
    title     = {Posture-Based Infant Action Recognition in the Wild With Very Limited Data},
    booktitle = {Proceedings of the IEEE/CVF Conference on Computer Vision and Pattern Recognition (CVPR) Workshops},
    month     = {June},
    year      = {2023},
    pages     = {4912-4921}
}

@InProceedings{Amraee_2026_CVPR,
    author    = {Amraee, Somaieh and Singh, Ashutosh and McCullough, Aston and Scheithauer, Mindy and Goodwin, Matthew and Ostadabbas, Sarah},
    title     = {Toward Automated Behavior Understanding in Autism: A Zero-Shot Vision-Language Model Approach},
    booktitle = {Proceedings of the IEEE/CVF Conference on Computer Vision and Pattern Recognition (CVPR) Workshops},
    month     = {June},
    year      = {2026},
    pages     = {8733-8743}
}

@article{Popoola2012,
    author = {Popoola, Oluwatoyin P. and Wang, Kejun},
    title = {Video-Based Abnormal Human Behavior Recognition—A Review},
    year = {2012},
    issue_date = {November 2012},
    publisher = {IEEE Press},
    volume = {42},
    number = {6},
    issn = {1094-6977},
    url = {https://doi.org/10.1109/TSMCC.2011.2178594},
    doi = {10.1109/TSMCC.2011.2178594},
    journal = {Trans. Sys. Man Cyber Part C},
    month = nov,
    pages = {865–878},
    numpages = {14}
}

@inproceedings{Robertson2005,
    author = {Robertson, Neil and Reid, Ian},
    title = {Behaviour Understanding in Video: A Combined Method},
    year = {2005},
    isbn = {076952334X01},
    publisher = {IEEE Computer Society},
    address = {USA},
    url = {https://doi.org/10.1109/ICCV.2005.47},
    doi = {10.1109/ICCV.2005.47},
    booktitle = {Proceedings of the Tenth IEEE International Conference on Computer Vision (ICCV'05) Volume 1 - Volume 01},
    pages = {808–815},
    numpages = {8},
    series = {ICCV '05}
}

@article{Barami2024,
  author  = {Barami, Tal and Manelis-Baram, Liora and Kaiser, Hadas and others},
  title   = {Automated Analysis of Stereotypical Movements in Videos of Children With Autism Spectrum Disorder},
  journal = {JAMA Network Open},
  year    = {2024},
  volume  = {7},
  number  = {9},
  pages   = {e2432851},
  doi     = {10.1001/jamanetworkopen.2024.32851}
}

@misc{li2026deterministicevidencelayervisionlanguage,
      title={A Deterministic Evidence Layer for Vision-Language Autism Screening from Naturalistic Home Video}, 
      author={Wenqi Li and Mindi Ruan and Chuanbo Hu and Shuo Wang and Xin Li},
      year={2026},
      eprint={2610.09217},
      archivePrefix={arXiv},
      primaryClass={cs.CV},
      url={https://arxiv.org/abs/2610.09217}, 
}

@misc{huang2025hmorelearninghumancentricmotion,
      title={H-MoRe: Learning Human-centric Motion Representation for Action Analysis}, 
      author={Zhanbo Huang and Xiaoming Liu and Yu Kong},
      year={2025},
      eprint={2504.10676},
      archivePrefix={arXiv},
      primaryClass={cs.CV},
      url={https://arxiv.org/abs/2504.10676}, 
}

@article{Newell01061999,
author = {Karl M. Newell and Thomas Incledon and James W. Bodfish and Robert L. Sprague},
title = {Variability of Stereotypic Body-Rocking in Adults With Mental Retardation},
journal = {American Journal on Mental Retardation},
volume = {104},
number = {3},
pages = {279--288},
year = {1999},
publisher = {Taylor \& Francis},
doi = {10.1352/0895-8017(1999)104<0279:VOSBIA>2.0.CO;2},
URL = { 
        https://www.tandfonline.com/doi/abs/10.1352/0895-8017%281999%29104%3C0279%3AVOSBIA%3E2.0.CO%3B2
},
eprint = {   
        https://www.tandfonline.com/doi/pdf/10.1352/0895-8017%281999%29104%3C0279%3AVOSBIA%3E2.0.CO%3B2
}
}

@article{Hall01112003,
author = {Scott Hall and Tracy Thorns and Chris Oliver},
title = {Structural and Environmental Characteristics of Stereotyped Behaviors},
journal = {American Journal on Mental Retardation},
volume = {108},
number = {6},
pages = {391--402},
year = {2003},
publisher = {Taylor \& Francis},
doi = {10.1352/0895-8017(2003)108<391:SAECOS>2.0.CO;2},
URL = {    
        https://www.tandfonline.com/doi/abs/10.1352/0895-8017%282003%29108%3C391%3ASAECOS%3E2.0.CO%3B2
},
eprint = {    
        https://www.tandfonline.com/doi/pdf/10.1352/0895-8017%282003%29108%3C391%3ASAECOS%3E2.0.CO%3B2
}
}

@article{Kennedy2000Functions,
author = {Kennedy, Craig H. and Meyer, Kim A. and Knowles, Tanya and Shukla, Smita},
title = {ANALYZING THE MULTIPLE FUNCTIONS OF STEREOTYPICAL BEHAVIOR FOR STUDENTS WITH AUTISM: IMPLICATIONS FOR ASSESSMENT AND TREATMENT},
journal = {Journal of Applied Behavior Analysis},
volume = {33},
number = {4},
pages = {559-571},
doi = {https://doi.org/10.1901/jaba.2000.33-559},
url = {https://onlinelibrary.wiley.com/doi/abs/10.1901/jaba.2000.33-559},
eprint = {https://onlinelibrary.wiley.com/doi/pdf/10.1901/jaba.2000.33-559},
year = {2000}
}

@article{Symons2005Development,
author = {Symons, F. J. and Sperry, L. A. and Dropik, P. L. and Bodfish, J. W.},
title = {The early development of stereotypy and self-injury: a review of research methods},
journal = {Journal of Intellectual Disability Research},
volume = {49},
number = {2},
pages = {144-158},
doi = {https://doi.org/10.1111/j.1365-2788.2004.00632.x},
url = {https://onlinelibrary.wiley.com/doi/abs/10.1111/j.1365-2788.2004.00632.x},
eprint = {https://onlinelibrary.wiley.com/doi/pdf/10.1111/j.1365-2788.2004.00632.x},
year = {2005}
}

@article{moss2009prevalence,
  title={The prevalence and phenomenology of repetitive behavior in genetic syndromes},
  author={Moss, Joanna and Oliver, Chris and Arron, Kate and Burbidge, Cheryl and Berg, Katy},
  journal={Journal of autism and developmental disorders},
  volume={39},
  number={4},
  pages={572--588},
  year={2009},
  publisher={Springer}
}

@article{RAPP2005548,
title = {Stereotypy II: a review of neurobiological interpretations and suggestions for an integration with behavioral methods},
journal = {Research in Developmental Disabilities},
volume = {26},
number = {6},
pages = {548-564},
year = {2005},
issn = {0891-4222},
doi = {https://doi.org/10.1016/j.ridd.2004.11.006},
url = {https://www.sciencedirect.com/science/article/pii/S0891422205000089},
author = {John T. Rapp and Timothy R. Vollmer}
}

@article{Melo2020Prevalence,
  author  = {Melo, Cl{\'a}udia and Ruano, Lu{\'i}s and Jorge, Joana and Pinto Ribeiro, Tiago and Oliveira, Guiomar and Azevedo, Lu{\'i}s and Temudo, Teresa},
  title   = {Prevalence and determinants of motor stereotypies in autism spectrum disorder: A systematic review and meta-analysis},
  journal = {Autism},
  volume  = {24},
  number  = {3},
  pages   = {569--590},
  year    = {2020},
  doi     = {10.1177/1362361319869118}
}

@article{Goodwin2011Automated,
  author  = {Goodwin, Matthew S. and Intille, Stephen S. and Albinali, Fahd and Velicer, Wayne F.},
  title   = {Automated detection of stereotypical motor movements},
  journal = {Journal of Autism and Developmental Disorders},
  volume  = {41},
  number  = {6},
  pages   = {770--782},
  year    = {2011},
  doi     = {10.1007/s10803-010-1102-z}
}

@inproceedings{Rajagopalan2013SSBD,
  author    = {Rajagopalan, Shyam Sundar and Dhall, Abhinav and Goecke, Roland},
  title     = {Self-Stimulatory Behaviours in the Wild for Autism Diagnosis},
  booktitle = {Proceedings of the IEEE International Conference on Computer Vision (ICCV) Workshops},
  year      = {2013},
  pages     = {755--761},
  doi       = {10.1109/ICCVW.2013.103}
}

@article{Gilchrist2018Automated,
  author  = {Gilchrist, Kristin H. and Hegarty-Craver, Meghan and Christian, Robert B. and Grego, Sonia and Kies, Ashley C. and Wheeler, Anne C.},
  title   = {Automated detection of repetitive motor behaviors as an outcome measurement in intellectual and developmental disabilities},
  journal = {Journal of Autism and Developmental Disorders},
  volume  = {48},
  number  = {5},
  pages   = {1458--1466},
  year    = {2018},
  doi     = {10.1007/s10803-017-3408-6}
}
